\documentclass[review,5p]{elsarticle}

\usepackage{lineno,hyperref}
\modulolinenumbers[5]

\journal{Neurocomputing}

\usepackage{amsmath}
\usepackage{amsthm}
\usepackage{amssymb}
\usepackage{mathtools}
\usepackage{multirow}
\usepackage{tikz}
\usepackage{xcolor}
\usetikzlibrary{arrows}
\usepackage{enumitem}
\usepackage{algorithm}
\usepackage{algorithmic}
\usepackage{hyperref}
\usepackage{bm}
\usepackage{graphicx}
\usepackage{subfigure}
\usepackage{caption}
\newcommand{\mat}[1]{\mathbf{#1}}
\newcommand{\vect}[1]{\mathbf{#1}}

\newtheorem{thm}{Theorem}[]
\newtheorem{thm1}{Theorem}[]
\newtheorem{lem}{Lemma}[]

\newtheorem{defn}{Definition}[]
\newtheorem{defn1}{Definition}[]

\usepackage{diagbox} 
\begin{document}

\begin{frontmatter}

\title{Memorized Sparse Backpropagation}

\author[address1]{Zhiyuan Zhang}
\ead{zzy1210@pku.edu.cn}
\author[address1]{Pengcheng Yang}
\ead{yang\_pc@pku.edu.cn}
\author[address1]{Xuancheng Ren}
\ead{renxc@pku.edu.cn}
\author[address2,address1]{Qi Su\corref{mycorrespondingauthor}}
\ead{sukia@pku.edu.cn}
\author[address1]{Xu Sun\corref{mycorrespondingauthor}}
\ead{xusun@pku.edu.cn}

\address[address1]{MOE Key Laboratory of Computational Linguistics, School of EECS, Peking University, Beijing 100871, China.}
\address[address2]{School of Foreign Languages, Peking University, Beijing 100871, China.}

\cortext[mycorrespondingauthor]{Corresponding authors.}

\begin{abstract}
Neural network learning is usually time-consuming since backpropagation needs to compute full gradients and backpropagate them across multiple layers. Despite its success of existing works in accelerating propagation through sparseness, the relevant theoretical characteristics remain under-researched and empirical studies found that they suffer from the loss of information contained in unpropagated gradients. To tackle these problems, this paper presents a unified sparse backpropagation framework and provides a detailed analysis of its theoretical characteristics. Analysis reveals that when applied to a multilayer perceptron, our framework essentially performs gradient descent using an estimated gradient similar enough to the true gradient, resulting in convergence in probability under certain conditions. Furthermore, a simple yet effective algorithm named \textbf{m}emorized \textbf{s}parse \textbf{b}ack\textbf{p}ropagation (MSBP) is proposed to remedy the problem of information loss by storing unpropagated gradients in memory for learning in the next steps. Experimental results demonstrate that the proposed MSBP is effective to alleviate the information loss in traditional sparse backpropagation while achieving comparable acceleration.
\end{abstract}


\end{frontmatter}



\section{Introduction}
Training neural networks tends to be time-consuming~\cite{jean2014using,poultney2007efficient,seide20141}, especially for architectures with a large number of learnable model parameters. An important reason why neural network learning is typically slow is that backpropagation requires the calculation of full gradients and updates all parameters in each learning step~\cite{sun2017meprop}. As deep networks with massive parameters become more prevalent, more and more efforts are devoted to accelerating the process of backpropagation. Among existing efforts, a prominent research line is sparse backpropagation~\cite{sun2017meprop,wei2017minimal,DBLP:journals/corr/abs-1806-00512}, which aims at sparsifying the full gradient vector to achieve significant savings on computational cost.

One effective solution for sparse backpropagation is top-$k$ sparseness, which only keeps $k$ elements with the largest magnitude in the gradient vector and backpropagates them across different layers. For instance, meProp~\cite{sun2017meprop} employs the top-$k$ sparseness to compute only a very small but critical portion of the gradient information and update corresponding model parameters for the linear transformation. Going a step further, \cite{wei2017minimal} implements the top-$k$ sparseness for backpropagation on convolutional neural networks. Experimental results demonstrate that these methods can achieve a significant acceleration of the backpropagation process. However, despite its success in saving computational cost, the top-$k$ sparseness for backpropagation still suffers from some intractable drawbacks, elaborated on as follows.

On the theoretical side, the theoretical characteristics of sparse backpropagation, especially for top-$k$ sparseness~\cite{sun2017training,sun2017meprop,wei2017minimal}, have not been fully explored. Most previous work focuses on illustrating empirical explanations, rather than providing theoretical guarantees. Towards filling this gap, we first present a unified sparse backpropagation framework, of which some existing work~\cite{sun2017meprop,wei2017minimal} can prove to be special cases. Furthermore, we analyze the theoretical characteristics of the proposed framework, which provides theoretical explanations for some related works~\cite{sun2017training,sun2017meprop,wei2017minimal}. The relevant analysis illustrates that when applied to a multilayer perceptron, the proposed framework essentially employs an estimated gradient similar enough to the true gradient to perform gradient descent, which leads to convergence under certain conditions.

On the empirical side, we find that top-$k$ sparseness for backpropagation tends to cause information loss contained in unpropagated gradients. Although it can propagate the most crucial gradient information by keeping only $k$ elements with the largest magnitude in the gradient vector, the unpropagated gradient may also contain substantial useful information. Such information loss usually results in some adverse effects like poor stability in model performance. The model performance means task-specific evaluation scores, such as accuracy on classification tasks here.
To remedy this, we propose \textbf{m}emorized \textbf{s}parse \textbf{b}ack\textbf{p}ropagation (MSBP), which stores unpropagated gradients in memory for the next step of learning while propagating a critical portion of the gradient information. Compared to the previous works~\cite{sun2017meprop,wei2017minimal}, the proposed MSBP is capable of alleviating the information loss with the memory mechanism, thus improving model performance significantly.
To sum up, the main contributions of this work are two-fold:

\begin{enumerate}

    \item We present a unified sparse backpropagation framework and prove that some existing methods~\cite{sun2017meprop,wei2017minimal} are special cases under this framework. In addition, the theoretical characteristics of the proposed framework are analyzed in detail to provide theoretical accounts for related work.
    
    \item We propose memorized sparse backpropagation, which aims at alleviating the information loss by storing unpropagated gradients in memory for the next step of learning. The experiments demonstrate that our approach is able to effectively alleviate information loss while achieving comparable acceleration.

\end{enumerate}

\section{Related Work}

When training neural networks, the gradient to be backpropagated is not necessarily the true gradient. Synthetic gradients method~\citep{Understanding-Synthetic-Gradients-and-Decoupled-Neural-Interfaces} allows layers to be trained without waiting for the true error gradient backpropagated from the previous layer. The Direct Feedback Alignment method~\citep{direct-feedback-alignment-Provides-Learning} suggests that the weights used for gradient calculation in backward propagation do not have to be symmetric with the weights used for forward calculation. Furthermore, by combining extremely sparse connections with feedback-alignment causes a small accuracy drop while reducing multiply-and-accumulate (MAC) operations and data transmission cost~\citep{Direct-Feedback-Alignment-with-Sparse-Connections}. The calculation of the gradient also allows for the locality. Local Propagation~\citep{Local-Propagation-in-Constraint-based-Neural-Network} and Alternating Direction
Method of Multipliers (ADMM)~\citep{Training-Neural-Networks-Without-Gradients} calculate gradients locally to avoid long dependencies among variable gradients and enable the parallelization of the training computations over the neural units.

Since estimated gradients, rather than true gradients, can be used for backpropagation, a prominent research line to accelerate backpropagation is sparse backpropagation. It accelerates neural network training, which tends to be time-consuming~\cite{jean2014using,poultney2007efficient,seide20141}, by sparsifying gradients in backpropagation. For instance, a hardware-oriented structural sparsifying method~\cite{DBLP:journals/corr/abs-1806-00512} is invented for LSTM, which enforces a fixed level of sparsity in the LSTM gate gradients, yielding block-based sparse gradient matrices. \cite{sun2017meprop} proposes meProp for linear transformation, which employs top-$k$ sparseness to computes only a small but critical portion of gradients and updates corresponding model parameters. Furthermore, mePorp can also be extended to convolutional layers~\citep{wei2017minimal} and deep sequence-to-sequence models~\cite{sun2017training}. 

Besides sparse backpropagation, sparse gradient methods calculate the true gradients in backpropagation but sparsify gradients for parameter updates or communication in a distributed system. Many efforts are devoted to analyzing the convergence of sparse gradient method~\citep{NIPS2018_7697,The-Convergence-of-Sparsified-Gradient-Methods,Variance-Reduction-with-Sparse-Gradients}. \cite{NIPS2018_7697} proposes to equip sparse gradient methods with an error memory in a distributed system.

Sparse coding~\cite{olshausen1996natural} is a kind of unsupervised methods to represent data or features efficiently and accelerate neural networks training, whose plausibility is tested in the literature biologically~\citep{Neural-correlates-of-sparse-coding-and-dimensionality-reduction,constraining-the-connectivity,Sparse-Coding-with-a-Somato-Dendritic-Rule}. The sparse auto-encoder~\cite{poultney2007efficient} learns sparse features with an energy-based model to represent them efficiently. In order to train outrageously large neural networks, \cite{shazeer2017outrageously} introduces Sparsely-Gated Mixture-of-Experts (MoE) layer. A gating network is utilized to select a sparse combination of the expert networks in MoE. Sparse representation has also been applied to computer vision problems, for example, \cite{DBLP:journals/tnn/ZhengLYTT18} proposes a sparse temporal encoding method to get visual features for robust object recognition.

There are also many approaches that do not utilize sparsity to accelerate network learning. For example, \cite{tallaneare1990fast} proposes an adaptive acceleration strategy for backpropagation while \cite{riedmiller1993direct} performs local adaptation of parameter update based on error function. To speed up the computation of the softmax layer, \cite{jean2014using} utilizes importance sampling to make the training more efficient. \cite{dropout} presents dropout, which improves training speed and reduces overfitting by randomly dropping units from the neural network during training. From the perspective of distributed systems, \cite{seide20141} proposes a one-bit-quantizing mechanism to reduce the communication cost between multiple machines.

\section{Preliminary}
\label{sec:pre}
This section presents some preliminary preparations.
Given the dataset $\mathcal{D}=\{(\vect{x},\vect{y})\}$, the training loss of an input instance $\vect{x}$ is defined as $\ell\big(\vect{w}; (\vect{x}, \vect{y})\big)$, where $\vect{w}$ denotes the learnable model parameters and $\ell(\cdot,\cdot)$ is some loss function such as $\ell_2$ or logistic loss. Further, the training loss on the whole dataset $\mathcal{D}$ is defined as:
\begin{equation}
\ell(\vect{w}; \mathcal{D})=\frac{1}{|\mathcal{D}|}\sum_{(\vect{x}, \vect{y})\in \mathcal{D}}\ell\big(\vect{w}; (\vect{x}, \vect{y})\big)
\end{equation}
We represent the angle between the vector $\vect{a}$ and the vector $\vect{b}$ as:
\begin{equation}
\angle \langle\vect{a}, \vect{b}\rangle=\arccos\frac{\vect{a}\bm{\cdot}\vect{b}}{\|\vect{a}\|\|\vect{b}\|} \in [0, \pi]
\end{equation}

For $\mu$-strongly convex\footnote{Suppose $\ell(\vect{w})=\ell(\vect{w};\mathcal{D})$, $\ell(\vect{w})$ is $\mu$-strongly convex if and only if for any vectors $\vect{a}, \vect{b}$, $\ell(\vect{b})\ge\ell(\vect{a})+\nabla\ell(\vect{a})\bm{\cdot}(\vect{b}-\vect{a})+\frac{\mu}{2}\|\vect{b}-\vect{a}\|^2$.} and $L$-smooth\footnote{Suppose $\ell(\vect{w})=\ell(\vect{w};\mathcal{D})$,  $\ell(\vect{w})$ is $L$-smooth if and only if for any vectors $\vect{a}, \vect{b}$, $ \ell(\vect{b})\le\ell(\vect{a})+\nabla\ell(\vect{a})\bm{\cdot}(\vect{b}-\vect{a})+\frac{L}{2}\|\vect{b}-\vect{a}\|^2$.} functions, $\mu/L$ often plays an important role in the convergence analysis~\cite{stich2018local,smooth-convex1,smooth-convex-condition}. \cite{smooth-convex-condition} uses the condition number $L/\mu$ to characterize the performance of their methods. We define the convex-smooth angle to characterize the convergence of our method: 

\begin{defn}[Convex-smooth angle]
\label{def:convex-smooth_angle}
If the training loss $\ell=\ell(\vect{w}; \mathcal{D})$ on the dataset $\mathcal{D}$ is $\mu$-strongly convex and $L$-smooth for parameter vector $\vect{w}$, the \textbf{convex-smooth angle} of $\ell$ is defined as:
\begin{equation}
\phi(\ell)=\arccos\sqrt{\frac{\mu}{L}}\in(0, \frac{\pi}{2})
\footnote{The condition $0<\mu<L$ always holds. Please refer to Appendix B.2 for the details.}
\end{equation}
\end{defn}

Then we define the gradient estimation angle to  measure estimation: 
\begin{defn}[Gradient estimation angle]
\label{def:estimation_angle}
For any vector $\vect{v}$ and training loss $\ell$ on an instance or whole dataset, we use $\vect{g^v}$ to represent an estimation of the true gradient $\frac{\partial \ell}{\partial\vect{v}}$. Then, the \textbf{gradient estimation angle} between the estimated gradient $\vect{g^v}$ and the true gradient $\frac{\partial \ell}{\partial\vect{v}}$ is defined as:
\begin{equation}
\delta(\vect{v})=\angle \langle\vect{g^v}, \frac{\partial \ell}{\partial \vect{v}}\rangle
\end{equation}
\end{defn}

The definition of sparsifying function is represented as:
\begin{defn}[Sparsifying function]
\label{def:sparsifing_function}
Given an integer $k\in[0, n]$, the function $S_{\mathbb{I}_k}(\cdot)$ is defined as $S_{\mathbb{I}_k}(\vect{v})=\mathbb{I}_k(\vect{v})\odot\vect{v}$,
where $\vect{v}\in\mathbb{R}^n$ is the input vector, and $\mathbb{I}_k(\vect{v})$ is a binary vector consisting of $k$ ones and $n-k$ zeros determined by $\vect{v}$. If $S_{\mathbb{I}_k}(\vect{v})$ satisfies that $\forall \vect{v}\in \mathbb{R}^n$:
\begin{equation}
    \langle S_{\mathbb{I}_k}(\vect{v}), \vect{v}\rangle\le\arccos\sqrt{\frac{k}{n}}
\end{equation}
we call $S_{\mathbb{I}_k}(\vect{v})$ \textbf{sparsifying function} and define its \textbf{sparse ratio} as $r=k/n$.
\end{defn}

\begin{defn}[$top_k$]
\label{def:topk_function}
Given an integer $k\in[0, n]$, for vector $\vect{v}=(v_1,\cdots,v_n)^{\rm T}\in\mathbb{R}^n$ where $|v_{\pi_1}|\ge\cdots\ge|v_{\pi_n}|$, the $top_k$ function is defined as:
\begin{equation}
top_k(\vect{v})=\mathbb{I}_k(\vect{v})\odot\vect{v}    
\end{equation}
where the $i$-th element of $\mathbb{I}_k(\vect{v})$ is $\mathbb{I}(i\in\{\pi_1,\cdots,\pi_k\})$. In other words, the $top_k$ function only preserves $k$ elements with the largest magnitude in the input vector.
\end{defn}

It is straightforward that $top_k$ is a special sparsifying function (see Appendix B.3).

\section{A Unified Sparse Backpropagation Framework}
\label{sec:usbp}
This section presents a unified framework for \textbf{s}parse \textbf{b}ack\textbf{p}ropagation (SBP), which can be used to explain some existing representative approaches~\cite{sun2017meprop,wei2017minimal}. 
We first define the \textbf{e}stimated \textbf{g}radient \textbf{d}escent (EGD) algorithm and then formally introduce the proposed framework.

\subsection{Estimated Gradient Descent}
\label{sec:egd}
Here we introduce the definition of the EGD algorithm, which serves as a base for analyzing the convergence of sparse backpropagation.

\begin{defn}[EGD]
Suppose $\ell=\ell(\vect{w}; \mathcal{D})$ is the training loss defined on the dataset $\mathcal{D}$ and $\vect{w} \in \mathbb{R}^n$ is the parameter vector to learn. The EGD algorithm adopts the following parameter update:
\begin{equation}
\vect{w}_{t+1}=\vect{w}_t-\eta_t \vect{g}_{t}^{\vect{w}}
\end{equation}
where $\vect{w}_t$ is the parameter at time-step $t$, $\eta_t > 0$ is the learning rate, and $\vect{g}_{t}^{\vect{w}}$ is an estimation of the true gradient $\frac{\partial\ell}{\partial\vect{w}_t}$ for parameter updates.
\end{defn}

Some existing optimizers can be regarded as special cases of EGD. For instance, when $\vect{g}_{t}^{\vect{w}}$ is defined as the true gradient $\frac{\partial\ell}{\partial\vect{w}_t}$, EGD is essentially the gradient descent (GD) algorithm. Several other works (e.g. Adam~\cite{DBLP:journals/corr/KingmaB14}, AdaDelta~\cite{zeiler2012adadelta}) can also be summarized as different expressions of EGD when $\vect{g}_{t}^{\vect{w}}$ is implemented as different estimates. 
More importantly, in essence, the sparse backpropagation employs the estimated gradient to approximate the true gradient for model training, which can also be regarded as a special case of EGD. This connection casts the cornerstone of subsequent theoretical analysis of sparse backpropagation.

In this work, we theoretically show that once the \emph{gradient estimation angle} $\delta(\vect{w}_t)$ of the parameter $\vect{w}_t$ satisfies certain conditions for each time-step $t$, the EGD algorithm can converge to the global minima $\vect{w}^*$ under some reasonable assumptions. This conclusion is demonstrated in Theorem~\ref{thm1}. Readers can refer to Appendix C.1 for the detailed proofs.


\begin{thm}[Convergence of EGD]
\label{thm1}
Suppose $\vect{w}_t$ is the parameter vector of time-step $t$, $\vect{w}^*$ is the global minima, and training loss $\ell=\ell(\vect{w}; \mathcal{D})$ defined on the dataset $\mathcal{D}$ is $\mu$-strongly convex and $L-$smooth.  When applying the EGD algorithm to minimize $\ell$, if the gradient estimation angle $\delta(\vect{w}_t)$ of $\vect{w}_t$ satisfies $\delta(\vect{w}_t)+\phi(\ell)\le \theta<{\pi}/{2}$, then there exists a learning rate $\eta_t>0$ for each time-step $t$ such that
\begin{equation}
\|\vect{w}_{t+1}-\vect{w}^*\|\le\sin\theta\|\vect{w}_{t}-\vect{w}^*\|
\end{equation}

Furthermore, $\vect{w}_{t}$ converges to $\vect{w}^*$ and the convergence speed is $O(\log\frac{1}{\epsilon})$:

$\forall \epsilon\in(0, \|\vect{w}_0-\vect{w}^*\|), \exists T(\epsilon)=\log\frac{ \|\vect{w}_0-\vect{w}^*\|}{\epsilon} \big/ \log{\frac{1}{\sin\theta}}$ s.t. $\forall T\ge T(\epsilon)$
\begin{equation}
\|\vect{w}_T-\vect{w}^*\| \le \epsilon
\end{equation}
where $T(\epsilon)$ is the maximum iteration number required for convergence (depending on $\varepsilon$). 
\end{thm}

For the given training loss $\ell=\ell(\vect{w}; \mathcal{D})$, $\phi(\ell)$ is a fixed value. Therefore, the Theorem~\ref{thm1} demonstrates that the EGD algorithm can converge to the global minima $\vect{w}^*$ and the convergence speed is $O(\log\frac{1}{\epsilon})$ when the gradient estimation angle $\delta(\vect{w}_t)$ between the estimated gradient $\vect{g}_{t}^{\vect{w}}$ and the true gradient $\frac{\partial\ell}{\partial\vect{w}_t}$ is small enough at each time-step. 

The insights gained from Theorem~\ref{thm1} can be generalized to non-convex loss functions, as evidenced in \cite{non-convex-review}. Notice that, even for non-convex loss functions $\ell$, there still exist neighbourhoods of every local minima $\vect{w}^*$, $\mathcal{A}\subset \mathbb{R}^n$, where the loss function is restricted $\mu$-strongly convex\footnote{$\ell(\vect{w})=\ell(\vect{w};\mathcal{D})$ is restricted $\mu$-strongly convex on $\mathcal{A}$ if and only if for any vectors $\vect{a}, \vect{b}\in \mathcal{A}$, $\ell(\vect{b})\ge\ell(\vect{a})+\nabla\ell(\vect{a})\bm{\cdot}(\vect{b}-\vect{a})+\frac{\mu}{2}\|\vect{b}-\vect{a}\|^2$.} and restricted $L$-smooth\footnote{$\ell(\vect{w})=\ell(\vect{w};\mathcal{D})$ is restricted $L$-smooth on $\mathcal{A}$ if and only if for any vectors $\vect{a}, \vect{b}\in \mathcal{A}$, $ \ell(\vect{b})\le\ell(\vect{a})+\nabla\ell(\vect{a})\bm{\cdot}(\vect{b}-\vect{a})+\frac{L}{2}\|\vect{b}-\vect{a}\|^2$.}. While the theoretical assumption of the loss function being strongly convex and smooth is partially true on $\mathcal{A}$ but not applicable to the entire $R^n$ in Theorem~\ref{thm1}, our results can be generalized to non-convex loss functions: when the weights $\vect{w}$ drop in $\mathcal{A} \subset \mathbb{R}^n$, it can converge to the local minima $\vect{w}^*$ in $\mathcal{A}$ and the convergence speed is $O(\log{1\over \epsilon})$.

Since the sparse backpropagation employs the estimated gradient to approximate the true gradient for model training, which can also be seen as a special case of EGD, Theorem~\ref{thm1} implies that gradient estimation angle can indicate the convergence of the sparse backpropagation algorithm.

\begin{figure}[ttt!]
\begin{minipage}[t]{0.49\textwidth}
\begin{algorithm}[H]
\small
\caption{Unified sparse backpropagation learning for a linear layer}
\label{code:sparse_bp}
\begin{algorithmic}[1]
   \STATE Initialize learnable parameter $\mat{W}$
   \STATE /* No memory here. */
   \WHILE{training}
       \STATE /* forward */
       \STATE Get input of this layer $\vect{x}$
       \STATE $\vect{h} \gets \mat{W}\vect{x}$
       \STATE $\vect{z} \gets \sigma(\vect{h})$
       \STATE Propagate $\vect{z}$ to the next layer
       \STATE /* backward */ 
       \STATE Get ${\partial \ell \over \partial \vect{z}}$ propagated  from the next layer 
       \STATE ${\partial \ell \over \partial \vect{h}} \gets \sigma'(\vect{h}) \odot {\partial \ell \over \partial \vect{z}}$
       \STATE $\vect{g} \gets S_{\mathbb{I}_k}({\partial \ell \over \partial \vect{h}})$
       \STATE /* Drop unpropagated part of ${\partial \ell \over \partial \vect{h}}$. */
       \STATE ${\partial \ell \over \partial \mat{W}} \gets \vect{g}\vect{x}^\text{T}$
       \STATE ${\partial \ell \over \partial \vect{x}} \gets \mat{W}^\text{T}\vect{g}$
       \STATE Backpropagate ${\partial \ell \over \partial \vect{x}}$ to the previous layer. 
       \STATE /* update */ 
       \STATE Update $\mat{W}$ with ${\partial \ell \over \partial \mat{W}}$
   \ENDWHILE
\end{algorithmic}
\end{algorithm}
\end{minipage}
\end{figure}

\subsection{Proposed Unified Sparse Backpropagation}
\label{sec:sparse_bp}
In this section, we present a unified sparse backpropagation framework via sparsifying function (Definition~\ref{def:sparsifing_function}). The core idea is that when performing backpropagation, the gradients propagated from the next layer are sparsified to achieve acceleration. Algorithm~\ref{code:sparse_bp} presents the pseudo-code of our unified sparse backpropagation framework, which is described in detail as follows.

Considering that a computation unit composed of one linear transformation and one activation function is the cornerstone of various neural networks, we elaborate on our unified sparse backpropagation framework based on such a computational unit:
\begin{align}
\vect{h} &= \mat{W}\vect{x}\\
\vect{z} &= \sigma(\vect{h})
\end{align}
where $\vect{x} \in \mathbb{R}^n$ is the input vector, $\mat{W} \in \mathbb{R}^{m\times n}$ is the parameter matrix, and $\sigma(\cdot):\mathbb{R}^m\to\mathbb{R}^m$ denotes a pointwise activation function. In MLP, if $\vect{x}$ represents the input of layer $l$, $\vect{z}$ can represent the output of layer $l$, which is also the input of layer $l+1$ if layer $l$ is not the last layer. Besides linear layers in MLP, the computation unit can also be fully-connected layers in CNN, gate layers in LSTM, etc.

Then, the original backpropagation computes the gradient of the parameter matrix $\mat{W}$ and the input vector $\vect{x}$ as follows:
\begin{align}
\frac{\partial\ell}{\partial \vect{h}}&=\frac{\partial\ell}{\partial \vect{z}}\odot{\sigma'(\vect{h})}\\
\frac{\partial\ell}{\partial \mat{W}}&=\frac{\partial\ell}{\partial \vect{h}}\vect{x}^\text{T}\\
\frac{\partial\ell}{\partial \vect{x}}&=\mat{W}^\text{T}\frac{\partial \ell}{\partial \vect{h}}
\end{align}

In the proposed unified sparse backpropagation framework, the sparsifying function (Definition~\ref{def:sparsifing_function}) is utilized to sparsify the gradient $\frac{\partial\ell}{\partial\vect{h}}$ propagated from the next layer and propagates them through the gradient computation graph according to the chain rule. Note that $\frac{\partial\ell}{\partial\vect{h}}$ is also an estimated gradient passed from the next layer. The gradient estimations are finally performed as follows:
\begin{align}
\frac{\partial\ell}{\partial \vect{h}}&\gets\frac{\partial\ell}{\partial \vect{z}}\odot{\sigma'(\vect{h})}\\
\frac{\partial\ell}{\partial \mat{W}}&\gets S_{\mathbb{I}_k}\left(\frac{\partial\ell}{\partial \vect{h}}\right)\vect{x}^\text{T}\\
\frac{\partial\ell}{\partial \vect{x}}&\gets\mat{W}^\text{T}S_{\mathbb{I}_k}\left(\frac{\partial \ell}{\partial \vect{h}}\right)
\end{align}

Since $top_k$ is a special sparsifying function (see Section~\ref{sec:pre}), some existing approaches (e.g., meProp~\cite{sun2017meprop}, meProp-CNN~\cite{wei2017minimal}) based on the top-$k$ sparseness can be regarded as special cases of our framework. Depending on the specific task, the sparsifying function can be defined as the different expressions to improve model performance. 

However, an intractable challenge for sparse backpropagation is the lack of theoretical analysis. To remedy this, here we analyze the theoretical characteristics of the proposed framework. 
With the fact that sparse backpropagation is a special case of EGD (Section~\ref{sec:egd}), we theoretically illustrate that when applied to a multi-layer perceptron (MLP), the proposed framework can converge to the global minima in probability under several reasonable conditions, which is formalized in Theorem~\ref{thm3}.

\begin{thm}[Gradient estimation angle of SBP]
\label{thm3}
Suppose:

(1) For the dataset dataset $\mathcal{D}$, $|\mathcal{D}|$, the size of dataset $\mathcal{D}$, is large enough and data instance $(\vect{x}, \vect{y})\in\mathcal{D}$ obeys independent and identical distribution (i.i.d.). 

(2) The neural network is an MLP model.\footnote{There are several trivial
constraints on MLP. Please refer to Appendix B.5 for more details.}

(3) We apply the unified sparse backpropagation to train the neural network and set the sparse ratio of every sparsifying function in SBP as $r$. 

Then we can get an estimation of  the training loss $\ell=\ell(\vect{w}; \mathcal{D})$ and $\delta(\vect{w})$, the gradient estimation angle of parameter vector $\vect{w}$, satisfies:

$\forall \theta\in (0, {\pi}/{2}), \exists r\in(0, 1)$,  s.t.
\begin{equation}
\lim\limits_{|\mathcal{D}|\to\infty}\mathrm{P}(\delta(\vect{w})<\theta)=1
\end{equation}
\end{thm}

The crucial idea to prove Theorem~\ref{thm3} is illustrating that the angle between the sparse gradient and the true full gradient can be small enough for every single data instance, and then prove that the gradient estimation angle of the full dataset can be small enough with probability one. Readers can refer to Appendix C.2 for the detailed proofs.

Theorem~\ref{thm3} reveals that the gradient estimation angle $\delta(\vect{w})$ of the parameter vector can be arbitrarily small with probability one when $\mathcal{D}$ is large enough. It implies that the sparse backpropagation algorithm is likely to converge because the gradient estimation angle can be small enough.

Although Theorem~\ref{thm3} is constrained by several additional conditions such as the base architecture of MLP, it is able to provide a degree of theoretical account for the proposed unified sparse backpropagation framework. Our efforts in these theoretical analyses are valuable because they help explain the effectiveness of not only our framework but also some existing approaches~\cite{sun2017meprop,wei2017minimal} on the theoretical side.

\begin{figure*}[!ht]
\centering
\footnotesize
\includegraphics[scale=0.4]{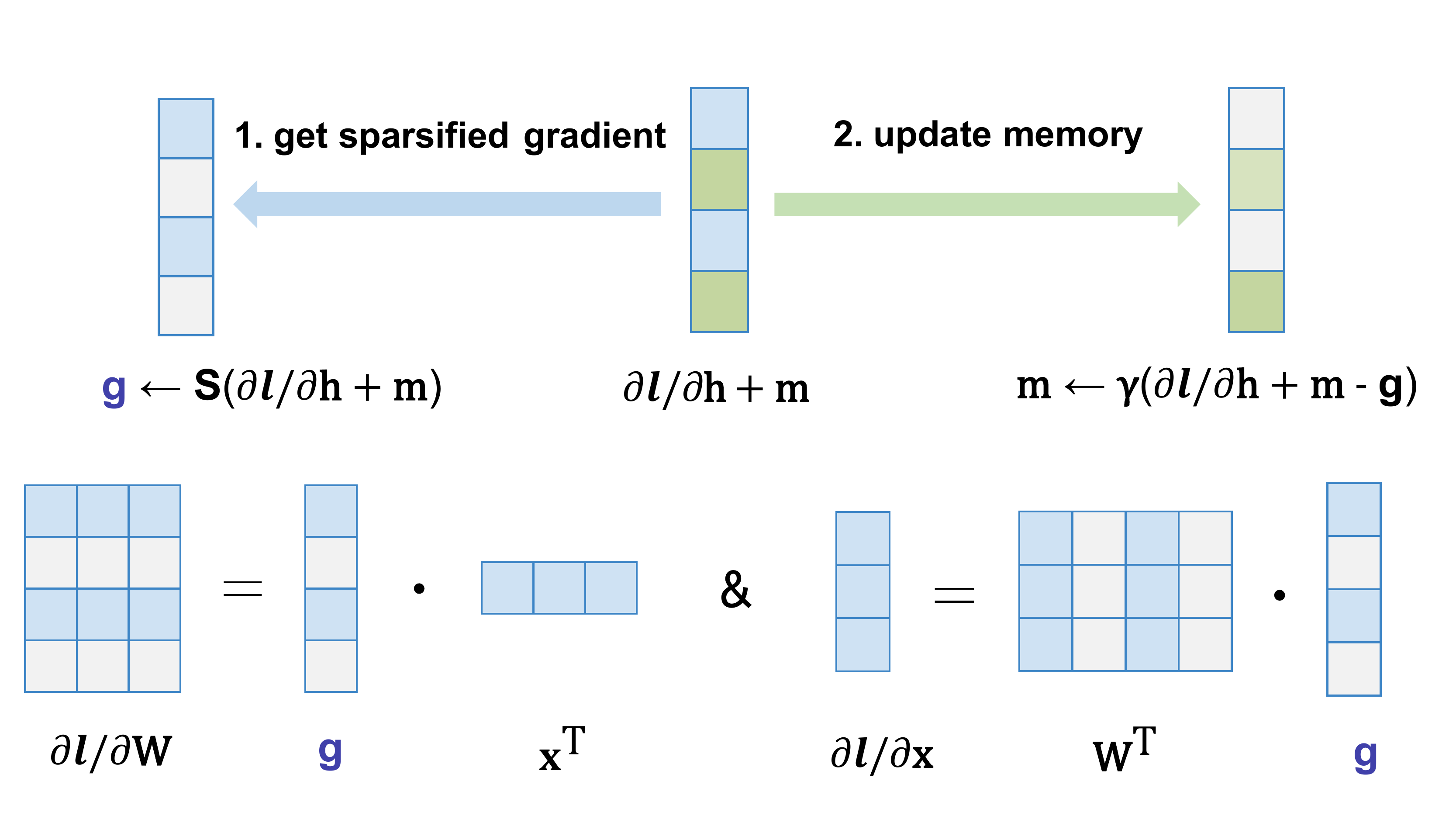}
\caption{An illustration of the dataflow of proposed MSBP method. Here $S$ denotes the sparsifying function.}
\label{fig:memory_dataflow}
\end{figure*}

\section{Memorized Sparse Backpropagation}
Although traditional sparse backpropagation is able to achieve significant acceleration by keeping only part of elements in the full gradient, the unpropagated gradient may also contain a certain amount of useful information. Experimental results find that such information loss tends to bring negative effects (e.g., performance degradation in extremely sparse scenarios, poor stability in performance). To remedy this, this work proposes \textbf{m}emorized \textbf{s}parse \textbf{b}ack\textbf{p}ropagation (MSBP), which aims at alleviating the information loss by storing unpropagated gradients in memory for the next step of learning

\subsection{Proposed Memory Sparse Backpropagation Method}
The core component of the proposed MSBP is the memory mechanism, which enables MSBP to store unpropagated gradients for the next step of learning while propagating a critical portion of the gradient information. Formally, different from the unified sparse backpropagation in Section~\ref{sec:sparse_bp}, we adopt the following gradient estimations:
\begin{align}
\frac{\partial\ell}{\partial \vect{h}}&\gets\frac{\partial\ell}{\partial \vect{z}}\odot{\sigma'(\vect{h})}\\
\frac{\partial\ell}{\partial \mat{W}}&\gets S_{\mathbb{I}_k}\left(\frac{\partial\ell}{\partial \vect{h}}+\vect{m}\right)\vect{x}^\text{T}\\
\frac{\partial\ell}{\partial \vect{x}}&\gets\mat{W}^\text{T}S_{\mathbb{I}_k}\left(\frac{\partial \ell}{\partial \vect{h}}+\vect{m}\right)
\end{align}
where $S_{\mathbb{I}_k}(\cdot)$ is a given sparsifying function and $\vect{m}$ is the memory storing unpropagated gradients from the last learning step. Then, the memory $\vect{m}$ is updated by the information of unpropagated gradients at the current learning step. Formally,
\begin{align}
\vect{m} \gets \gamma \left({\partial\ell \over \partial\vect{h}} + \vect{m} - S_{\mathbb{I}_k}\left({\partial \ell \over \partial \vect{h}}+\vect{m}\right)\right)
\end{align}
where $\gamma\in(0,1)$ is the memory ratio, a hyper-parameter controlling the ratio of memorizing unpropagated gradients. When $\gamma$ is set to 0, the proposed MSBP degenerates to the unified sparse backpropagation that completely discard unpropagated gradients. Before the model training begins, we initialize memory $\vect{m}$ to zero vector. Algorithm~\ref{code:memorized_sparse_bp} presents the pseudo code of MSBP. Figure~\ref{fig:memory_dataflow} presents the dataflow of the proposed MSBP.

\begin{figure}[ttt!]
\begin{minipage}[t]{0.49\textwidth}
\begin{algorithm}[H]
\small
\caption{Memorized sparse backpropagation learning for a linear layer}
\label{code:memorized_sparse_bp}
\begin{algorithmic}[1]
    \STATE Initialize learnable parameter$\mat{W}$ 
    \STATE Initialize gradient memory $\vect{m} \gets \textbf{0}$
   \WHILE{training}
       \STATE /* forward */ 
       \STATE Get input of this layer $\vect{x}$
       \STATE $\vect{h} \gets \mat{W}\vect{x}$
       \STATE $\vect{z} \gets \sigma(\vect{h})$
       \STATE Propagate $\vect{z}$ to the next layer
       \STATE /* backward */ 
       \STATE Get ${\partial \ell \over \partial \vect{z}}$ propagated  from the next layer 
       \STATE ${\partial \ell \over \partial \vect{h}} \gets \sigma'(\vect{h}) \odot {\partial \ell \over \partial \vect{z}}$
        \STATE $\vect{g} \gets S_{\mathbb{I}_k}({\partial \ell \over \partial \vect{h}}+\vect{m})$
        \STATE $\vect{m} \gets \gamma ({\partial \ell \over \partial \vect{h}} + \vect{m} -  \vect{g})$
       \STATE ${\partial \ell \over \partial \mat{W}} \gets \vect{g}\vect{x}^\text{T}$
       \STATE ${\partial \ell \over \partial \vect{x}} \gets \mat{W}^\text{T}\vect{g}$
       \STATE Backpropagate ${\partial \ell \over \partial \vect{x}}$ to the previous layer
       \STATE /* update */ 
       \STATE Update $\mat{W}$ with ${\partial \ell \over \partial \mat{W}}$
   \ENDWHILE
\end{algorithmic}
\end{algorithm}
\end{minipage}
\vskip 0in
\end{figure}

Intuitively, by storing unpropagated gradients with the memory mechanism, the information loss in backpropagation due to sparseness can be alleviated. The experiments also illustrate that the proposed MSBP is more advantageous in various respects than approaches that completely discards unpropagated gradients. 
In fact, we find that for MSBP, the angle between the sparse gradient and true full gradient tends to be small.
Furthermore, this angle is smaller than that in traditional sparse backpropagation. 
According to theoretical analysis in Section~\ref{sec:usbp}, a smaller \emph{gradient estimation angle} is more conducive to model convergence. 
This observation explains the effectiveness of our MSBP to a certain extent on the theoretical side. Readers can refer to Section~\ref{analysis} for a more detailed analysis.

\textbf{Comparison to sparsified SGD with memory.}
A work that looks similar to this paper is sparsified SGD with memory~\cite{DBLP:conf/nips/StichCJ18}, which equips sparse gradients with memory. 
It calculates full gradients in backpropagation and sparsifies them to be communicated in a distributed system. However, the backpropagation process remains unchanged and cannot be accelerated. Different from sparsified SGD with memory, we sparsify gradients in backpropagation and both the communication and backpropagation process can be accelerated.
Besides, sparsified SGD with memory is an optimization approach that can only be used in distributed systems, while our MSBP can be applied to both distributed and centralized systems.

\subsection{Implementations}

Following meProp~\cite{sun2017meprop}, we adopt $S_{\mathbb{I}_k}(\cdot)=top_k(\cdot)$ as the sparsifying function. 

For simplicity, we use SBP to represent the traditional \textbf{s}parse \textbf{b}ack\textbf{p}ropagation that completely discards unpropagated gradients with $top_k(\cdot)$ sparsifying function and MSBP to represent the proposed \textbf{m}emorized \textbf{s}parse \textbf{b}ack\textbf{p}ropagation with $top_k(\cdot)$ sparsifying function.

\subsection{Discussion of Complexity Information}
\label{sec:Complexity} 
Table~\ref{Complexity_Table} presents a comparison of the time and memory complexity of traditional SBP and our proposed MSBP. In this section, we discuss the time complexity and memory complexity of traditional SBP and our proposed MSBP.
 
\textbf{Time complexity.}
The backpropagation process of the linear layer focuses on calculating gradients of $\mat{W}$ and $\vect{x}$, the time complexity of which is $O(mn)$. The application of SBP consists of two steps: finding top-$k$ dimensions of the gradient of $\vect{h}$ using a maximum heap with time complexity of $O(n \log k)$ and backpropagating only top-$k$ dimensions of gradients with time complexity of $O(mk)$. 
The extra time cost of MSBP comes from adding the memory information into the gradient of $\vect{h}$ and updating the memory. The time complexity of these two operations is $O(n)$, which is negligible compared to $O(mk+n \log k)$.

\textbf{Memory complexity.}
The analysis of memory complexity is similar.
The backpropagation of the linear layer requires storing gradients of $\mat{W}$ and $\vect{x}$, whose memory complexity is $O(mn)$. 
For traditional SBP, the memory complexity of finding top-$k$ dimensions of the gradient of $\vect{h}$ with a maximum heap is $O(k)$, while the backpropagation of corresponding dimensions of gradients requires no additional memory overhead. The extra memory cost of MSBP is the memory vector, the memory complexities of which are both $O(n)$ and negligible compared to $O(mn)$.

\begin{table}
\caption{The time and memory complexity of backpropagation for a linear layer with input size $n$ and output size $m$. We adopt $top_k$ as the sparsifying function.}
\label{Complexity_Table}
\footnotesize
\setlength{\tabcolsep}{6.0pt}
\centering
\begin{tabular}{|l|c|c|c|}
\hline
\textbf{Method} & Time & Memory\\
\hline
Linear&$O(mn)$&$O(mn)$\\
\hline
+ SBP&$O(mk+n\log k)$&$O(mn)$\\
+ MSBP&$O(mk+n\log k)$&$O(mn)$\\
\hline
\end{tabular}
\end{table}

\section{Experiments}

\begin{table*}[!t]
\footnotesize
\caption{Results of time cost and evaluation scores. $h$, $r$ and $\gamma$ refer to the hidden size, sparse ratio and memory ratio of our models \textit{(SBP, MSBP share the same $h$ with baseline and MSBP shares the same $r$ with SBP)}.
\textbf{BP (s)} and \textbf{Total (s)} refer to the backpropagation time cost and the total time cost on CPU in seconds \textit{($a \times$ is compared to baseline)}.
\textbf{Acc (\%)} and \textbf{UAS (\%)} refer to the averaged accuracy and unlabeled attachment score, respectively \textit{($\pm a$ is compared to baseline)}.}
\centering
\begin{tabular}{|l|c|c|c|}
\hline
\textbf{MNIST} & \textbf{BP (s)} & \textbf{Total (s)} & \textbf{Acc (\%)} \\
\hline
MLP ($h$=500) &67.2 (1.00$\times$)& 116.6 (1.00$\times$) &97.86 (+0.00)\\
\hline
+ SBP ($r$=0.04) &6.6 (10.18$\times$)&54.5 (2.14$\times$)&97.84 (-0.02)\\
+ MSBP ($\gamma$=0.8) & 6.9 (9.74$\times$)&55.4 (2.10$\times$) &\textbf{98.23 (+0.37)}\\
\hline\hline
\textbf{Parsing} & \textbf{BP (s)}& \textbf{Total (s)} & \textbf{UAS (\%)}\\
\hline
MLP ($h$=500) &6447 (1.00$\times$) &9016 (1.00$\times$)& 88.38 (+0.00)\\
\hline
+ SBP ($r$=0.04) &682 (9.46$\times$) &2886 (3.12$\times$)& 88.59 (+0.21)\\
+ MSBP ($\gamma$=0.7) &684 (9.43$\times$) & 2898 (3.11$\times$) &\textbf{89.03 (+0.65)}\\
\hline\hline
\textbf{POS-Tag}  & \textbf{BP (s)} & \textbf{Total (s)}& \textbf{Acc (\%)} \\
\hline
LSTM ($h$=500) &11965 (1.00$\times$) &16052 (1.00$\times$)&{97.27 (+0.00)}\\
\hline
+ SBP ($r$=0.04) &1763 (6.79$\times$)&5738 (2.80$\times$)&{97.34 (+0.07)}\\
+ MSBP ($\gamma$=0.8) &1842 (6.50$\times$)&5849 (2.74$\times$)&{\textbf{97.50 (+0.23)}}\\
\hline
\end{tabular}
\label{Memory1}
\end{table*}

\subsection{Experimental Settings}

We evaluate the proposed MSBP on several typical benchmark tasks. The baselines used for comparison on each task are also introduced.

\textbf{MNIST image recognition (MNIST).}
MNIST handwritten digit dataset~\cite{lecun1998gradient} aims to recognize the numerical digit (0-9) of each image. 
The numbers of training, development, and test images are 55,000, 5000, and 10,000 respectively. 
The evaluation metric is classification accuracy. 
The base model is a 3-layer MLP.

\textbf{Transition-based dependency parsing (Parsing).}
In this task, we use English Penn TreeBank (PTB)~\cite{DBLP:journals/coling/MarcusSM94} for experiments. 
The numbers of training, development and test transitions are $1,900,056$, $80,234$ and $113,368$ respectively. Each transition example contains a parsing context and its optimal transition action.
The evaluation metric is the unlabeled attachment score (UAS). 
Following~\cite{DBLP:conf/emnlp/ChenM14}, the base model is an MLP-based parser. In training, gradients are clipped~\cite{pascanu2013difficulty} to 5. 

\textbf{Part-of-speech tagging (POS-Tag).}
In this task, we use the standard benchmark dataset derived from Penn Treebank corpus~\cite{DBLP:conf/emnlp/Collins02}.
The numbers of training and test examples are $38,219$ and $5,462$ respectively. 
The evaluation metric is per-word accuracy.
Following~\cite{bilstm-for-tagging}, the base model is a $2$-layer bi-directional LSTM (Bi-LSTM). In addition, we use $100$-dim pre-trained GloVe~\cite{DBLP:conf/emnlp/PenningtonSM14} embeddings to initialize the word embeddings.

The epochs during training are set as $20, 20$, and $10$ on the three tasks of MNIST, Paring, and POS-Tag respectively. The dropout~\cite{dropout} probability is set to $0.1, 0.2, 0.5$ respectively. The Adam optimizer~\cite{DBLP:journals/corr/KingmaB14} with a learning rate of $10^{-3}$ is used on all three tasks. The batch size is set to $32, 1024$, and $128$, respectively. The experiments of time cost are conducted on a computer with the configuration of Intel(R) Core(TM) i5-8400 CPU @ 2.80 GHz CPU. 

Besides MNIST, Parsing, and POS Tagging tasks, we also conduct experiments on CNN and sequence-to-sequence models to illustrate the universality of our proposed method to a wide range of network architectures.

\textbf{Polarity classification (Polarity) and subjectivity classification (Subjectivity).}
The dataset is constructed by~\cite{meprop_cnn_dataset}. Both tasks are designed to perform sentence classification, with accuracy as the evaluation metric. For these two tasks, every experiment is repeated for $10$ times and report the averaged accuracy on the test set. 
The base model is TextCNN~\cite{kim2014convolutional}. The filter window sizes of TextCNN are $3$, $4$, and $5$, with $100$ feature maps each. The optimizer is Adam and the learning rate is $10^{-3}$. The batch size is set to $32$. We train models for $10$ epochs.

\textbf{English-Vietnamese Translation (En-Vi).}
The translated TED talks of IWSLT 2015 Evaluation Campaign~\cite{en_vi}, containing $133K$ training sentence pairs, are adopted as the training data. TED tst2012 and TED tst2013 are adopted as the development set and test set respectively.

\textbf{Simplified Chinese-English Translation (Chs-En).}
Following~\cite{cns_en}, LDC simplified Chinese-English Translation dataset, containing $1.25M$ sentence pairs with about $28M$ Chinese words and about $35M$ English words, is adopted as the training data. NIST 2002 and NIST 2003-2006 are adopted as the development set and the test set respectively. The test set is merged by NIST 2003-2006.

On En-Vi and Chs-En translation tasks, the evaluation metric is BLEU score~\cite{bleu}.
The base model is LSTM-based sequence-to-sequence (seq-to-seq) model. The encoder is a $3$-layer bidirectional LSTM encoder and the decoder is a $3$-layer LSTM decoder. The dropout rate is $0.4$ and $0.3$ respectively. The embedding size and hidden size are both $512$. The attention type is Luong-style~\cite{luong} and the beam
search size is $10$~\cite{beam}.
The optimizer is Adam and the learning rate is $10^{-3}$. The mini-batch size is set to $64$. The epochs during training are set as $40$ and $80$ epochs respectively on translation tasks.
SBP and MSBP are applied to each hidden layer on base models.

\subsection{Experimental Results}
\label{experimental_result}

The experimental results on three tasks of MNIST, Parsing, and POS-Tag are shown in Table~\ref{Memory1}. An in-depth analysis of the results is provided from the following aspects.

\textbf{Improving model performance.}
As shown in Table~\ref{Memory1}, the proposed MSBP achieves the best performance on all tasks. 
Considering that our ultimate goal is to accelerate neural network learning while achieving comparable model performance, such results are promising and gratifying. Compared to traditional SBP~\cite{sun2017meprop,wei2017minimal}, MSBP employs the memory mechanism to store unpropagated gradients. This reduces the information loss during backpropagation, leading to improvements in the model performance.

\textbf{Accelerating backpropagation.}
In contrast to traditional SBP, our MSBP memorizes unpropagated gradients to alleviate information loss. However, a potential issue is that the introduction of memory containing unpropagated gradients may impair the acceleration of backpropagation. 
As shown in Table~\ref{Memory1}, either traditional SBP or our proposed MSBP is able to achieve great acceleration of backpropagation, and the latter shows an only negligible increase in computational cost compared to the former. 
This illustrates that our MBSP can achieve comparable acceleration while improving model performance.

\begin{table}[t]
\footnotesize
\caption{Results of different approaches on TextCNN. \textbf{Acc} denotes the averaged accuracy.}
\centering
\begin{tabular}{|l|c|}
\hline
\textbf{Subjectivity} & \textbf{Acc (\%)}\\
\hline
 TextCNN & 93.66 (+0.00) \\
\hline
+ SBP ($r$=0.05) & 93.77 (+0.11)\\
+ MSBP ($r$=0.05, $\gamma$=0.6) & \textbf{93.80 (+0.14)}\\
\hline\hline
\textbf{Polarity} & \textbf{Acc (\%)} \\
\hline
 TextCNN & 80.89 (+0.00) \\
\hline
+ SBP ($r$=0.05) & 81.12 (+0.23) \\
+ MSBP ($r$=0.05, $\gamma$=0.3) & \textbf{81.48 (+0.58)}\\
\hline
\end{tabular}
\label{meProp-CNN}
\end{table}

\begin{table}[t]
\footnotesize
\caption{Results of different approaches on Sequence-to-Sequence models. \textbf{Dev BLEU} and \textbf{Test BLEU} refer to the BLEU score on development set and test set.}
\centering
\begin{tabular}{|l|c|c|}
\hline
\textbf{En-Vi} & \textbf{Dev BLEU} & \textbf{Test BLEU}\\
\hline
Seq2seq ($h$=512) & 25.87 & 28.45 (+0.00)\\
\hline
+ SBP ($r$=1/16) & 26.64 & 28.94 (+0.39)\\
+ MSBP ($r$=1/16, $\gamma$=0.05)  & 26.53 & \textbf{29.10 (+0.65)}\\
\hline\hline
\textbf{Chs-En} & \textbf{Dev BLEU} & \textbf{Test BLEU}\\
\hline
Seq2seq ($h$=512) & 38.54  & 35.86 (+0.00)\\
\hline
+ SBP ($r$=1/16) & 38.74 & 35.94 (+0.08)\\
+ MSBP ($r$=1/16, $\gamma$=0.05) & 38.28 & \textbf{35.98 (+0.12)}\\
\hline
\end{tabular}
\label{meProp-seq2seq}
\end{table}

\textbf{Applying to CNNs and Deep Seq-to-seq Models.}
The SBP and MSBP methods are also applicable to convolution layers and deep sequence-to-sequence models. Following mePorop-CNN~\cite{wei2017minimal}, The SBP method is implemented on CNNs, and then MSBP on CNNs, which is similar to the SBP method. Following alternating Top-k selection~\cite{sun2017training}, the sequence-to-sequence version of SBP, the SBP, and MSBP methods are applied to the encoder and the decoder iteratively.
As shown in Table~\ref{meProp-CNN} and \ref{meProp-seq2seq}, the proposed MSBP outperforms the baseline and the SBP method on TextCNN and deep deep sequence-to-sequence models.

\subsection{Related Systems of Evaluation Tasks}

This section presents evaluation scores of related systems on each task to illustrate the competitive performance of our approach. To testify the effectiveness of the proposed method, more advanced deep learning models are implemented, including the MLP, LSTM, CNN, and sequence-to-sequence models.

For MLP, the MLP based approaches can achieve around $98\%$~\cite{cirecsan2010deep,lecun1998gradient} accuracy on MNIST, while our method achieves $98.23\%$. For LSTM, the reported accuracy in existing approaches lies between $97.2\%$ to $97.4\%$~\cite{collobert2011natural,huang2015bidirectional,tsuruoka2011learning} on POS-Tag, whereas our method can achieve $97.50\%$ accuracy. As for CNN models, TextCNN~\cite{kim2014convolutional} reports around $81.3\%$ and $93.4\%$ on polarity classification and subjectivity classification respectively, while our method achieves around $81.5\%$ and $93.8\%$ respectively.

\section{Further In-Depth Analysis}
\label{analysis}

\subsection{Influence of different hyper-parameters.}

\begin{figure}[!t]
\centering
\footnotesize
\begin{minipage}[]{0.48\linewidth}  
\includegraphics[scale=0.2]{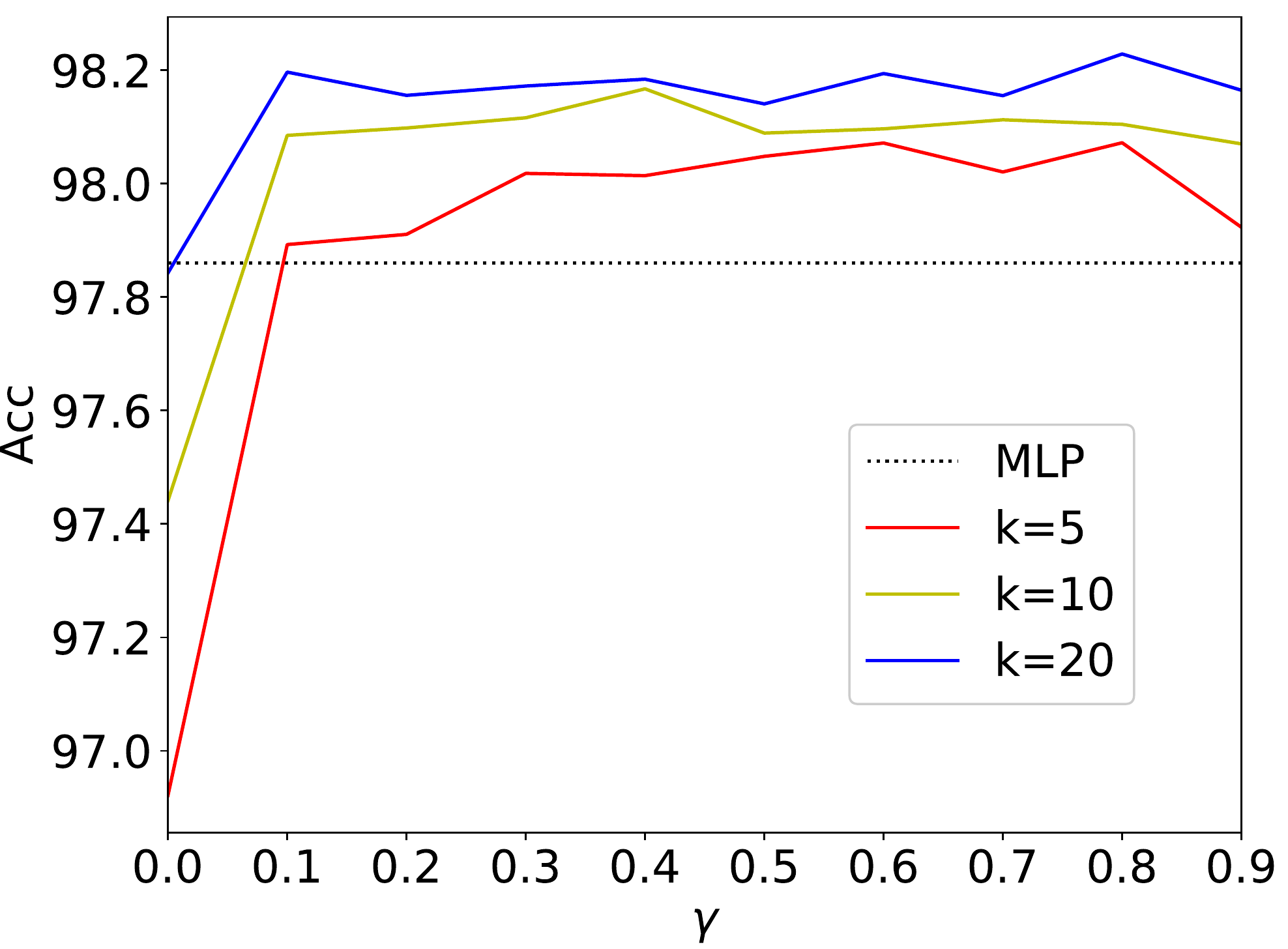}
\caption{Average accuracy.}
\label{acc}
\end{minipage}
\hfill
\begin{minipage}[]{0.48\linewidth}
\includegraphics[scale=0.2]{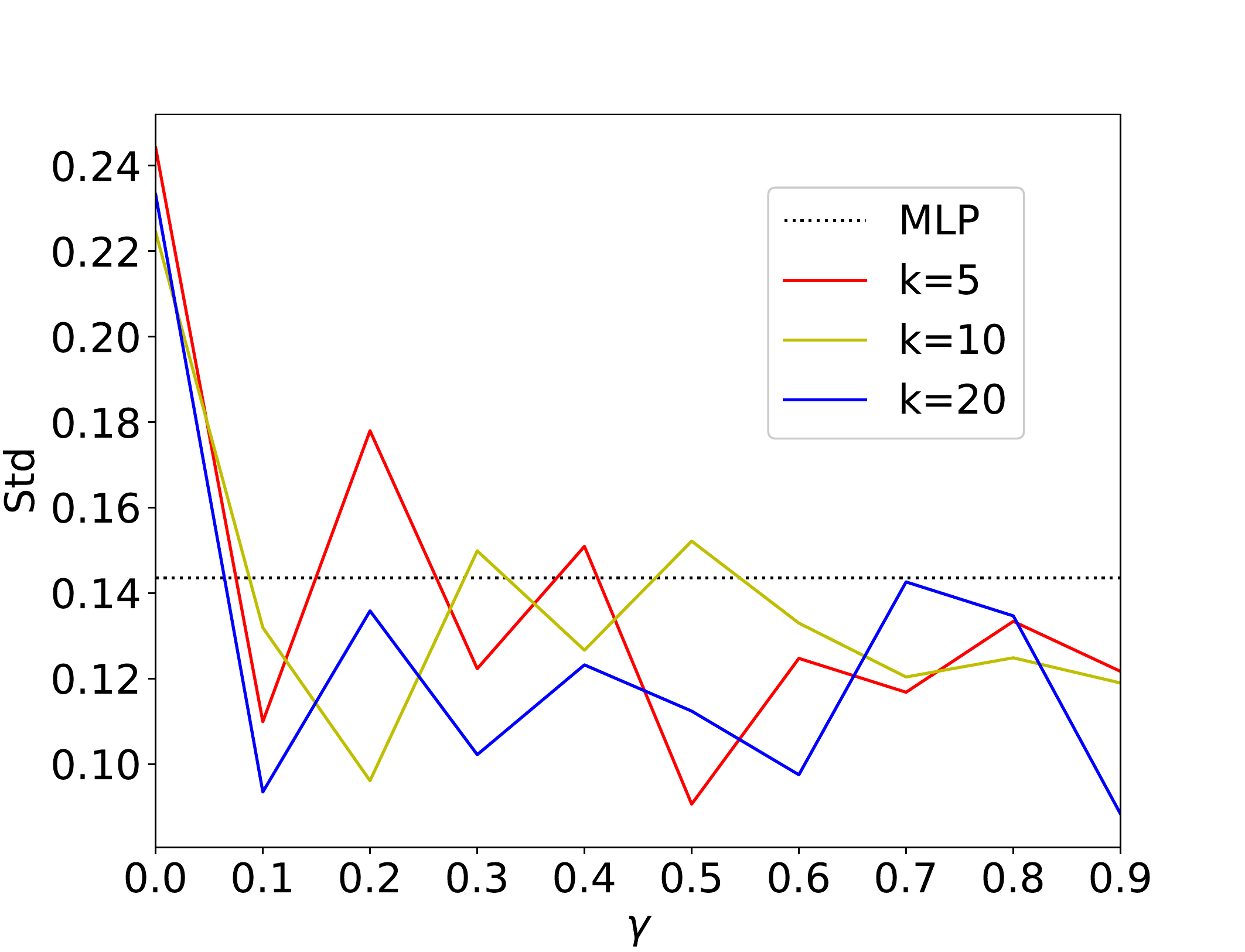}
\caption{Standard deviation.}
\label{std}
\end{minipage}
\end{figure}

\begin{figure}[!t]
\centering
\footnotesize
\begin{minipage}[]{0.48\linewidth}
\includegraphics[scale=0.26]{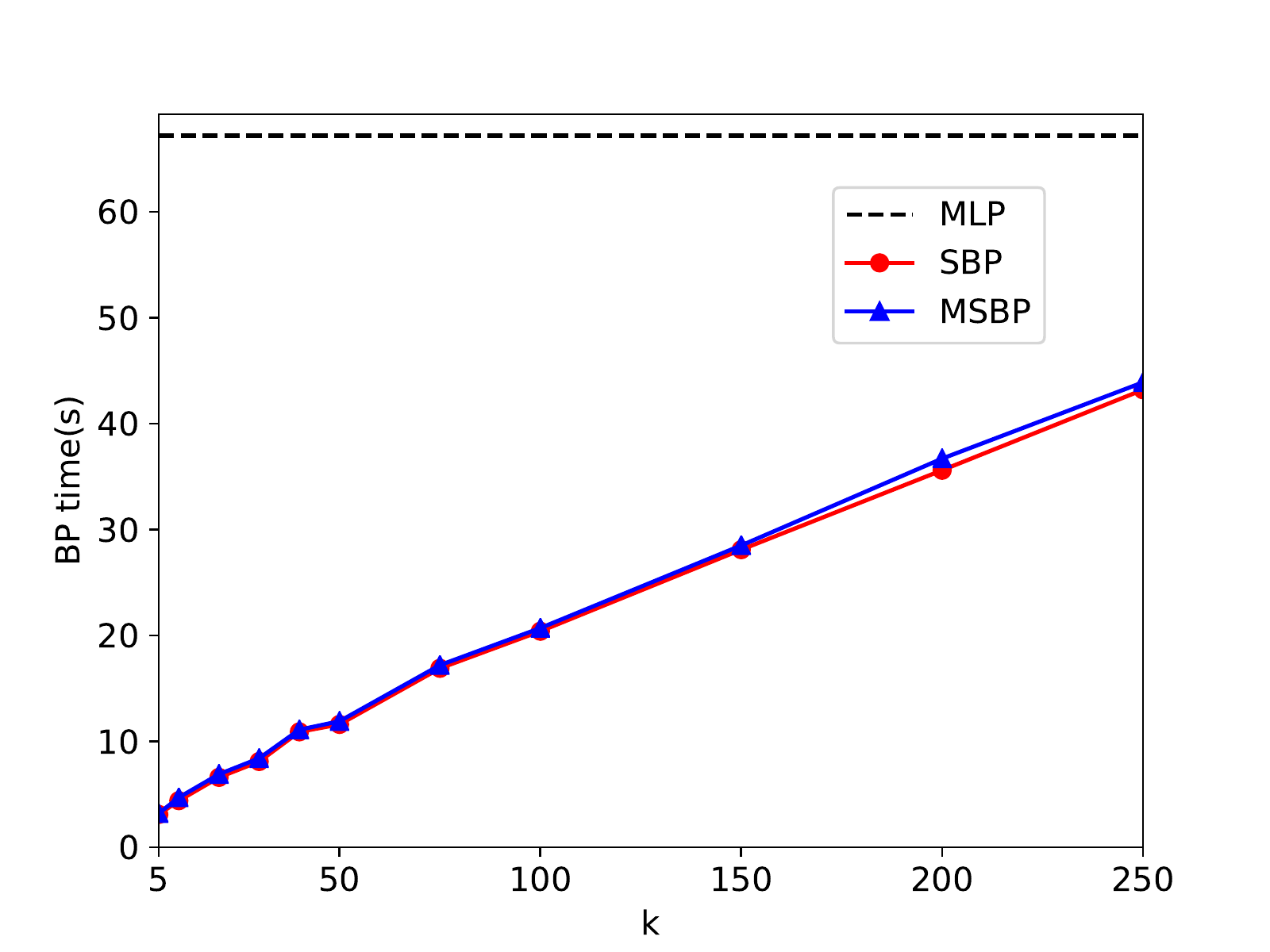}
\caption{BP time.}
\label{bptime}
\end{minipage}
\hfill
\begin{minipage}[]{0.48\linewidth}  
\includegraphics[scale=0.26]{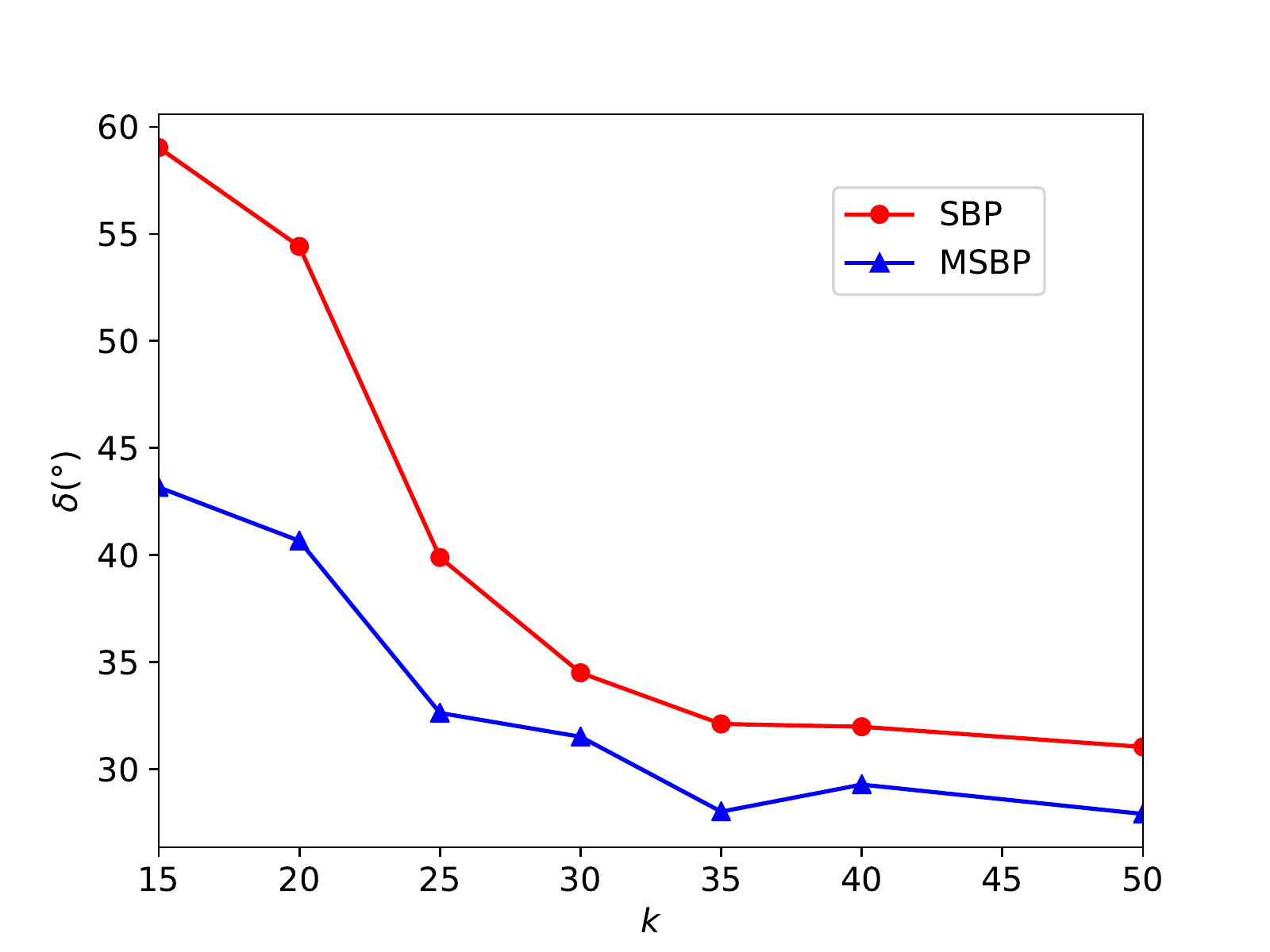}
\caption{Estimation angle.}
\label{angle}
\end{minipage}
\end{figure}

In order to explore the influence of different hyper-parameters on model performance and stability, for experiments on MNIST dataset, we select the sparse ratio $k$ among $\{5, 10, 20\}$ and the memory ratio $\gamma$ among $\{0.1, 0.2, 0.3, \cdots, 0.9\}$. All experiments are repeated 20 times for each setup of $k$ and $\gamma$. The mean and standard deviation of the accuracy of repeated experiments are presented in Figure~\ref{acc} and Figure~\ref{std}, respectively. The influence of hyper-parameters on backpropagation time cost is also explored, the results are shown in Figure~\ref{bptime}.

\textbf{Model performance.}
As depicted in Figure~\ref{acc}, smaller $k$ tends to lead to a worse performance both for both SBP and MSBP because only a small amount of gradient information is propagated when $k$ is small and the backpropagation suffers from information loss. For the same $k$, MSBP ($\gamma>0$) performs better than traditional SBP ($\gamma=0$) regardless of the choice of $\gamma$. The difference in accuracy between MSBP and traditional SBP ranges from 0.4\% to 0.8\%, while that between MSBP with different $\gamma$ settings lies between 0.1\% and 0.3\%. This implies that the performance of MSBP is not very sensitive to $\gamma$ compared to the improvement gained by the memory mechanism.

\textbf{Model stability.}
Our proposed MSBP also has higher stability than SBP, indicating that it contributes to reducing the variance of the model performance in repeated experiments. Figure~\ref{std} shows that the traditional SBP ($\gamma=0$) suffers from poor model stability in repeated experiments, whose standard deviation is nearly $1.7$ times of the base model (MLP). In contrast, all experiments conducted with MSBP ($\gamma>0$) have higher stability than traditional SBP regardless of the choice of $\gamma$.

\textbf{Backpropagation time cost.}
In experiments, the forward propagation time costs of the base model and our proposed SBP and MSBP models are nearly the same. Therefore, we analyze the backpropagation time cost.
Figure~\ref{bptime} shows the backpropagation time cost with different settings ($\gamma=0.1$). In experiments, choices of $\gamma$ do not influence the time cost of MSBP significantly.
As analyzed in Section~\ref{sec:Complexity}, the extra time cost of MSBP is negligible compared to SBP, which is also verified in Figure~\ref{bptime}.
In Figure~\ref{bptime}, backpropagation time is approximately proportional to $k$ (or the sparse ratio $r$), which accords our analysis in Section~\ref{sec:Complexity}. Higher $k$ tends to lead to better performance but less backpropagation acceleration. It means there exists a tradeoff between backpropagation acceleration and the performance of SBP and MSBP.

\subsection{Further verification.}

\textbf{Analysis of run-time gradient estimation angles.}
We further verify whether our proposed MSBP can give a more accurate estimation of the true gradient. As analyzed in Section~\ref{sec:usbp}, for sparse backpropagation, a smaller \emph{gradient estimation angle} can better guarantee the convergence of approach. Therefore, we calculate the \emph{gradient estimation angle} of the average gradients on the whole dataset in the proposed MSBP and traditional SBP to empirically explain the effectiveness of our method ($\gamma=0.1$). As shown in Figure~\ref{angle}, higher $k$ results in smaller \emph{gradient estimation angles} and for the same $k$, the \emph{gradient estimation angles} in MSBP are smaller than that in SBP. This illustrates that by employing the memory mechanism to store unpropagated gradients, the sparse gradient calculated by our approach gives a more accurate estimation of the true gradient, which is also consistent with results in Figure~\ref{acc}. In addition, the gap between the \emph{gradient estimation angles} of SBP and MSBP tends to be bigger for lower $k$ because SBP suffers from the loss of unpropagated gradients more for lower $k$, under which circumstances our proposal improves the performance to a larger extent. 

\textbf{Statistical test.}
Further, we conduct statistical tests to verify that: for MSBP and SBP with the same sparse ratio, 1) MSBP outperforms traditional SBP under different settings; and 2) MSBP has better stability than traditional SBP under different settings. The results show that for nearly all settings of $k$ and $\gamma$, MSBP has both better performance and stability than traditional SBP statistically significantly $(p<0.05)$. Please refer to Appendix D for details.

\begin{table}[!t]
\tiny
\caption{Results of time cost and evaluation scores on extremely sparse scenarios.}
\centering
\renewcommand\tabcolsep{3pt}
\begin{tabular}{|p{85pt}|p{44pt}|p{44pt}|p{46pt}|}
\hline
\textbf{MNIST} & \textbf{BP (s)} & \textbf{Total (s)} & \textbf{Acc (\%)} \\
\hline
MLP($h$=5000)&3667.5 (1.00$\times$)&5744.1 (1.00$\times$)&98.10(+0.00)\\
\hline
+ SBP ($r$=0.001)&57.0 (64.34$\times$)&2131.8 (2.71$\times$)&96.19(-1.93)\\
+ MSBP ($r$=0.001, $\gamma$=0.8) &58.8 (62.37$\times$) &2139.0 (2.70$\times$)& \textbf{97.71(-0.39)}\\
\hline
+ SBP ($r$=0.002) & 102.1 (35.92$\times$)&2196.4 (2.63$\times$)&96.22 (-1.90)\\
+ MSBP ($r$=0.002, $\gamma$=0.8) &104.7 (35.03$\times$) & 2206.8 (2.62$\times$)&\textbf{98.16 (+0.06)}\\
\hline
\end{tabular}
\label{sparse}
\end{table}

\subsection{Applicability to multiple scenarios}

\textbf{Applicability to extremely sparse scenarios.}
In sparse backpropagation, the sparse ratio $r$ controls the trade-off between acceleration and model performance. In pursuit of ultra-large acceleration, $r$ tends to be set extremely small values in real-life scenarios.
However, we empirically find that traditional SBP usually results in a significant degradation in model performance in this case.
Table~\ref{sparse} shows that traditional SBP brings a $1.90\%$ or $1.93\%$ reduction in accuracy on MNIST image classification for $r=0.001$ or $r=0.002$, which is a notable gap and ruins the practicality of applying meProp in extremely sparse cases. The reason is that for small $r$ values, only a very small amount of gradient information is propagated. Therefore, there exists serious information loss during backpropagation, leading to a significant degradation in model performance.
In contrast, results show that in these extremely sparse scenarios, the loss of the accuracy of MSBP compared to the baseline is tolerable when $r=0.001$ and our MSBP works as effectively as the base model when $r=0.002$. 
With the memory mechanism, the current unpropagated gradient information is stored for the next step of learning, reducing the information loss caused by sparseness. 

\textbf{Applicability to different base network architectures.}
We compare traditional SBP and the proposed MSBP on the other base network architectures, which is the CNN-based model and deep sequence-to-sequence models, to verify the universality of our approach in Section~\ref{experimental_result}. Experimental results in Table~\ref{meProp-CNN} and \ref{meProp-seq2seq} show that the proposed MSBP improves the performance of the base model on both sentence classification and machine translation tasks. This demonstrates that our MSBP is universal, which applies to different types of base networks and tasks.

\textbf{Applicability to large-scale datasets.} Experiments are also conducted on simplified Chinese-English translation task, which requires a large-scale dataset containing $1.25M$ sentence pairs. Experimental results in Table~\ref{meProp-seq2seq} demonstrate that the proposed MSBP can be also applied to the large-scale dataset.

\section{Conclusion and Future Work}
This work presents a unified \textbf{s}parse \textbf{b}ack\textbf{p}ropagation (SBP) framework. Some previous representative approaches can be regarded as special cases under this framework. Besides, the theoretical characteristics of the proposed framework are analyzed in detail to provide theoretical accounts for the relevant methods. Analysis reveals that when applied to a multilayer perceptron, our framework essentially performs gradient descent using an estimated gradient similar enough to the true gradient, resulting in convergence in probability under certain conditions. Going a step further, we propose \textbf{m}emorized \textbf{s}parse \textbf{b}ack\textbf{p}ropagation (MSBP), which aims at alleviating the information loss in tradition sparse backpropagation by utilizing the memory mechanism to store unpropagated gradients. The experiments demonstrate that the proposed MSBP exhibits better performance while achieving comparable acceleration. Further analysis also shows that the performance of MSBP is not very sensitive to memory ratio compared to the improvement gained by the memory mechanism. The proposed MSBP method also has higher stability than the SBP method and the extra time cost of MSBP is negligible compared to the SBP method.

In this work, the memory ratio is set as an adjustable hyper-parameter. However, the structural characteristics of different samples show natural differences, which may make the optimal memory ratio vary. Therefore, we will study adaptive methods that can automatically control the ratio of memorizing unpropagated gradients in the future.

\bibliography{main}

\onecolumn 
\appendix

\section{Review of Definitions and Theorems in Paper}
In this section, we review some important definitions and theorems introduced in the paper.

\subsection{Definitions}Given the dataset $\mathcal{D}=\{(\vect{x},\vect{y})\}$, the training loss of an input instance $\vect{x}$ is defined as $\ell\big(\vect{w}; (\vect{x}, \vect{y})\big)$, where $\vect{w}$ denotes the learnable model parameters and $\ell(\cdot,\cdot)$ is some loss function such as $\ell_2$ or logistic loss. Further, the training loss on the whole dataset $\mathcal{D}$ is defined as:
\begin{equation}
\ell(\vect{w}; \mathcal{D})=\frac{1}{|\mathcal{D}|}\sum_{(\vect{x}, \vect{y})\in \mathcal{D}}\ell\big(\vect{w}; (\vect{x}, \vect{y})\big)
\end{equation}
We represent the angle between the vector $\vect{a}$ and the vector $\vect{b}$ as:
\begin{equation}
\angle \langle\vect{a}, \vect{b}\rangle=\arccos\frac{\vect{a}\bm{\cdot}\vect{b}}{\|\vect{a}\|\|\vect{b}\|} \in [0, \pi].
\end{equation}

\begin{defn1}[Convex-smooth angle]
\label{def:convex-smooth_angle}
If the training loss $\ell=\ell(\vect{w}; \mathcal{D})$ on the dataset $\mathcal{D}$ is $\mu$-strongly convex and $L$-smooth for parameter vector $\vect{w}$, the \textbf{convex-smooth angle} of $\ell$ is defined as:
\begin{equation}
\phi(\ell)=\arccos\sqrt{\frac{\mu}{L}}\in(0, \frac{\pi}{2})
\footnote{The condition $0<\mu<L$ always holds. Please refer to Appendix.B.2 for the details.}
\end{equation}

\end{defn1}

\begin{defn1}[Gradient estimation angle]
\label{def:estimation_angle}
For any vector $\vect{v}$ and training loss $\ell$ on an instance or whole dataset, we use $\vect{g^v}$ to represent an estimation of the true gradient $\frac{\partial \ell}{\partial\vect{v}}$. Then, the \textbf{gradient estimation angle} between the estimated gradient $\vect{g^v}$ and the true gradient $\frac{\partial \ell}{\partial\vect{v}}$ is defined as:
\begin{equation}
\delta(\vect{v})=\angle \langle\vect{g^v}, \frac{\partial \ell}{\partial \vect{v}}\rangle\footnote{After both the training loss $\ell$ and the estimation method of the gradient are defined, the \emph{gradient estimation angle} $ \delta(\vect{v})$ depends only on $\vect{v}$.}
\end{equation}
\end{defn1}

\begin{defn1}[Sparsifying function]
\label{def:sparsifing_function}
Given an integer $k\in[0, n]$, the function $S_{\mathbb{I}_k}(\cdot)$ is defined as $S_{\mathbb{I}_k}(\vect{v})=\mathbb{I}_k(\vect{v})\odot\vect{v}$,
where $\vect{v}\in\mathbb{R}^n$ is the input vector, and $\mathbb{I}_k(\vect{v})$ is a binary vector consisting of $k$ ones and $n-k$ zeros determined by $\vect{v}$. If $S_{\mathbb{I}_k}(\vect{v})$ satisfies that $\forall \vect{v}\in \mathbb{R}^n$:
\begin{equation}
    \langle S_{\mathbb{I}_k}(\vect{v}), \vect{v}\rangle\le\arccos\sqrt{\frac{k}{n}}
\end{equation}
we call $S_{\mathbb{I}_k}(\vect{v})$ \textbf{sparsifying function} and define its \textbf{sparse ratio} as $r=k/n$.
\end{defn1}

\begin{defn1}[$top_k$]
\label{def:topk_function}
Given an integer $k\in[0, n]$, for vector $\vect{v}=(v_1,\cdots,v_n)^{\rm T}\in\mathbb{R}^n$ where $|v_{\pi_1}|\ge\cdots\ge|v_{\pi_n}|$, the $top_k$ function is defined as:
\begin{equation}
top_k(\vect{v})=\mathbb{I}_k(\vect{v})\odot\vect{v}    
\end{equation}
where the $i$-th element of $\mathbb{I}_k(\vect{v})$ is $\mathbb{I}(i\in\{\pi_1,\cdots,\pi_k\})$. In other words, the $top_k$ function only preserves $k$ elements with the largest magnitude in the input vector.
\end{defn1}

It is easy to verify that $top_k$ is a special sparsifying function (see Appendix.B.3.).

\begin{defn1}[EGD]
Suppose $\ell=\ell(\vect{w}; \mathcal{D})$ is the training loss defined on the dataset $\mathcal{D}$ and $\vect{w} \in \mathbb{R}^n$ is the parameter vector to learn. The \textbf{estimated gradient descent} (EGD) algorithm adopts the following parameter update:
\begin{equation}
\vect{w}_{t+1}=\vect{w}_t-\eta_t \vect{g}_{t}^{\vect{w}}
\end{equation}
where $\vect{w}_t$ is the parameter at time-step $t$, $\eta_t > 0$ is the learning rate, and $\vect{g}_{t}^{\vect{w}}$ is an estimation of the true gradient $\frac{\partial\ell}{\partial\vect{w}_t}$ for parameter updates.
\end{defn1}

\subsection{Theorems}


\begin{thm1}[Convergence of EGD]
\label{thm1}
Suppose $\vect{w}_t$ is the parameter vector of time-step $t$, $\vect{w}^*$ is the global minima, and training loss $\ell=\ell(\vect{w}; \mathcal{D})$ defined on the dataset $\mathcal{D}$ is $\mu$-strongly convex and $L-$smooth.  When applying the EGD algorithm to minimize $\ell$, if the gradient estimation angle $\delta(\vect{w}_t)$ of $\vect{w}_t$ satisfies $\delta(\vect{w}_t)+\phi(\ell)\le \theta<{\pi}/{2}$, then there exists learning rate $\eta_t>0$ for each time-step $t$ such that
\begin{equation}
\|\vect{w}_{t+1}-\vect{w}^*\|\le\sin\theta\|\vect{w}_{t}-\vect{w}^*\|
\end{equation}

Furthermore, $\vect{w}_{t}$ converges to $\vect{w}^*$ and the convergence speed is $O(\log\frac{1}{\epsilon})$:

$\forall \epsilon\in(0, \|\vect{w}_0-\vect{w}^*\|), \exists T(\epsilon)=\log\frac{ \|\vect{w}_0-\vect{w}^*\|}{\epsilon} \big/ \log{\frac{1}{\sin\theta}}$ s.t. $\forall T\ge T(\epsilon)$
\begin{equation}
\|\vect{w}_T-\vect{w}^*\| \le \epsilon
\end{equation}
where $T(\epsilon)$ is the maximum iteration number required for convergence (depending on $\varepsilon$). 
\end{thm1}

\begin{thm1}[Bounded gradient estimation angle of SBP]
\label{thm3}
Suppose:

(1) For the dataset dataset $\mathcal{D}$, $|\mathcal{D}|$, the size of dataset $\mathcal{D}$, is large enough and data instance $(\vect{x}, \vect{y})\in\mathcal{D}$ obeys independent and identical distribution (i.i.d.). 

(2) The neural network is a MLP model.

(3) We apply the unified sparse backpropagation to train the neural network and set the sparse ratio of every sparsifying function in SBP as $r$. 

Then we can get an estimation of  the training loss $\ell=\ell(\vect{w}; \mathcal{D})$ and $\delta(\vect{w})$, the gradient estimation angle of parameter vector $v$, satisfies:

$\forall \theta\in (0, {\pi}/{2}), \exists r\in(0, 1)$, s.t.
\begin{equation}
\lim\limits_{|\mathcal{D}|\to\infty}\mathrm{P}(\delta(\vect{w})<\theta)=1
\end{equation}
\end{thm1}


\section{Preparation and Lemmas}
Here we introduce some key definitions and lemmas throughout the appendix. 
All vectors and matrices are assumed to belong to the real number field. 
In Appendix, vectors (e.g. $\vect{x}, \vect{y}$) and matrices (e.g. $\mat{W}, \vect{A}$) are in bold formatting.

\subsection{Vectors}
We first introduce two vector-related lemmas.

\begin{lem}
\label{lemma.vector.1}
For any vectors $\vect{a}$, $\vect{b}$ and $\vect{c}$, we have
$$\angle\langle\vect{a},\vect{b}\rangle \le \angle\langle\vect{a},\vect{c}\rangle + \angle\langle\vect{b},\vect{c}\rangle$$
\end{lem}

\begin{lem}
\label{lemma.vector.2}
For matrix $\mat{A}\in \mathbb{R}^{m\times n}$ ($m
\ge n$), suppose $\mat{A}\mat{A}^\text{T}$ is a positive definite matrix, the eigenvalue decomposition of $\mat{A}\mat{A}^\text{T}$ is $\mat{A}\mat{A}^\text{T}=\mat{P}\mat{\Sigma} \mat{P}^\text{T}\in \mathbb{R}^{m\times m}$, $\mat{\Sigma}=\text{diag}\{\sigma_1, \sigma_2, \cdots\, \sigma_m \}$ ($\sigma_i>0$) and $\mat{P}$ is an orthogonal matrix. We define ${\sigma_\text{min}}=\min\limits_{i}\sigma_i$ and  ${\sigma_\text{max}}=\max\limits_{i}\sigma_i$. If $\rho\ge\frac{\sigma_\text{max}}{\sigma_\text{min}}\ge 1$, for any $n$-dimension vectors $\vect{u}$ and $\vect{v}$, we have
$$\cos\angle\langle \mat{A}^\text{T}u, \mat{A}^\text{T}v\rangle\ge\rho\cos\angle\langle u, v\rangle+1-\rho$$
\end{lem}

\subsection{Loss Function}
We define the loss function $\ell=\ell(\vect{w}; \mathcal{D})$ as $\mu$-strongly convex
if for $\mu>0, \nabla^2 \ell(\vect{x}) \succeq \mu I$,  where $I$ denotes identity matrix. If the loss function $\ell$ is $\mu$-strongly convex , for any vectors $\vect{a}, \vect{b}$, we have
\begin{align} 
\|\nabla\ell(\vect{a})-\nabla\ell(\vect{b})\|\ge\mu\|\vect{a}-\vect{b}\|  \label{mu.1}\\
\ell(\vect{b})\ge\ell(\vect{a})+\nabla\ell(\vect{a})\bm{\cdot}(\vect{b}-\vect{a})+\frac{\mu}{2}\|\vect{b}-\vect{a}\|^2
\label{mu.2}
\end{align}

We define the loss function $\ell=\ell(\vect{w}; \mathcal{D})$ as $L$-smooth
if for $L>0,  \nabla^2 \ell(\vect{x}) \preceq LI$, where $I$ denotes identity matrix. If the loss function $\ell$ is $L$-smooth, for any vectors $\vect{a}, \vect{b}$, we have
\begin{align}
\|\nabla\ell(\vect{a})-\nabla\ell(\vect{b})\|\le L\|\vect{a}-\vect{b}\| \label{L.1}\\
\ell(\vect{b})\le\ell(\vect{a})+\nabla\ell(\vect{a})\bm{\cdot}(\vect{b}-\vect{a})+\frac{L}{2}\|\vect{b}-\vect{a}\|^2 \label{L.2}
\end{align}

For the loss function $\ell$, we define $\vect{w}^*$ as its global minima. If $\ell$ is $L$-smooth, for any $\vect{w}$, we have
\begin{align}
\ell(\vect{w}^*) &\le \ell\big(\vect{w}-{1\over L}\nabla \ell(\vect{w})\big) \quad \text{(Because $\vect{w}^*$ is the global minima)}\\
&\le \ell(\vect{w})-\nabla\ell(\vect{w}) \bm{\cdot}{1\over L}\nabla \ell(\vect{w})+{L\over 2}\|-{1\over L}\nabla \ell(\vect{w})\|^2 \quad \text{(Ineq.(\ref{L.2}))}\\
&=  \ell(\vect{w})-{1\over 2L}||\nabla \ell(\vect{w})||^2 \label{equ-min}
\end{align}

From Eq.(\ref{L.1}) and Eq.(\ref{L.2}), we can see, 
\begin{align}
\ell(\vect{a})+\nabla\ell(\vect{a})\bm{\cdot}(\vect{b}-\vect{a})+\frac{\mu}{2}\|\vect{b}-\vect{a}\|^2
\le\ell(\vect{b})
\le\ell(\vect{a})+\nabla\ell(\vect{a})\bm{\cdot}(\vect{b}-\vect{a})+\frac{L}{2}\|\vect{b}-\vect{a}\|^2
\end{align}
In other words, $\mu\le L$. When $\mu=L$, we have $\ell(\vect{b})=\ell(\vect{a})+\nabla\ell(\vect{a})\bm{\cdot}(\vect{b}-\vect{a})+\frac{L}{2}\|\vect{b}-\vect{a}\|^2$. When we set $\vect{a}=\vect{0}=(0, 0,\cdots,0)^\text{T}$ and $\vect{b}=\vect{x}$, $\ell(\vect{x})=\ell(\vect{0})+\ell(\vect{a})\bm{\cdot}\vect{x}+\frac{L}{2}\|\vect{x}\|^2$, where it has a closed-form solution and is trival. Therefore, we assume in most cases, $0<\mu<L$. 

Back to the definition of convex-smooth angle, if the loss function $\ell$ is $\mu$-strongly convex and $L$-smooth ($0<\mu<L$), we can see the convex-smooth angle of $\ell$ is $\phi(\ell)=\arccos\sqrt{{\mu}/{L}}\in(0, \frac{\pi}{2})$.

\subsection{$top_k$ Function}
We will prove the $top_k$ function is a special sparsifying function.

Given an integer $k\in[0,n]$, for vector $\vect{v}=(v_1,\cdots,v_n)^{\rm T}\in\mathbb{R}^n$ where $|v_{\pi_1}|\ge\cdots\ge|v_{\pi_n}|$, the $top_k$ function is defined as $top_k(\vect{v})=\mathbb{I}_k(\vect{v})\odot\vect{v}$ where the $i$-th element of $\mathbb{I}_k(\vect{v})$ is $s_i=\mathbb{I}(i\in\{\pi_1,\cdots,\pi_k\})$. It is easy to verify that 
\begin{align}
\cos\angle\langle top_k(\vect{v}), \vect{v}\rangle 
=\frac{top_k(\vect{v})\bm{\cdot} \vect{v}}{\|top_k(\vect{v})\|\|\vect{v}\|} 
=\frac{\sum\limits_{i=1}^n(s_i v_i^2)}{\|{topk}(\vect{v})\|\|\vect{v}\|} 
=\frac{\sum\limits_{i=1}^n(s_i v_i)^2}{\|{topk}(\vect{v})\|\|\vect{v}\|} \\
=\frac{\|{topk}(\vect{v})\|^2}{\|{topk}(\vect{v})\|\|\vect{v}\|}
=\frac{\|{topk}(\vect{v})\|}{\|\vect{v}\|}
=\sqrt{\frac{\sum\limits_{i=1}^kv_{\pi_i}^2}{{\sum\limits_{i=1}^nv_{\pi_i}^2}}}
\ge\sqrt{\frac{k}{n}}
\end{align}

Therefore, the $top_k$ function is a  special sparsifying function.

\subsection{Linear Layer Trained with SBP}
Consider a linear layer with one linear transformation and one increasing pointwise activation function
\begin{align}
\vect{h} = \mat{W}\vect{x},\quad
\vect{z} = \sigma(\vect{h})
\end{align}
where $\vect{x} \in \mathbb{R}^n$ is the input sample, $\mat{W} \in \mathbb{R}^{m\times n}$ is the parameter matrix ($m \ge n$), $n$ is the dimension of the input vector, $m$ is the dimension of the output vector and $\sigma$ is an increasing pointwise activation function (e.g., $\sigma(x)=x$, $\sigma(x)=\tanh(x)$ or $\sigma(x)=\text{sigmoid}(x)$).

For matrix $\mat{W}\in\mathbb{R}^{m\times n}$, we define flattening function to flatten it into a vector in $\mathbb{R}^{nm}$ as $flatten(\mat{W}) = [\mat{W}_{:,1}; \cdots; \mat{W}_{:,n}]$, where $\mat{W}_{:,i}$ represents the $i$-th column of $\mat{W}$ and the semicolon denotes the concatenation of many column vectors to a long column vector. In other words, $flatten(\mat{W})_{(j-1)m+i} = W_{ij}$.

Assume $\vect{x}=(x_1, x_2, \cdots, x_n)^\text{T}, \vect{h}=(h_1, h_2, \cdots, h_m)^\text{T}$ and $\mat{W}=(W_{ij})_{1\le i\le m, 1\le j\le n}$, then $h_i=\sum\limits_{j=1}^nW_{ij}x_j$, when backpropagating
\begin{align}
\frac{\partial\ell}{\partial W_{ij}}=\frac{\partial\ell}{\partial h_i}{x_j},\quad
\frac{\partial\ell}{\partial x_{i}}=\sum\limits_{j=1}^m\frac{\partial\ell}{\partial h_j}{W_{ji}},\quad
\frac{\partial\ell}{\partial h_{i}}=\frac{\partial\ell}{\partial z_i}{\sigma'(h_i)}
\end{align}

Assume $\frac{\partial\ell}{\partial \vect{x}}=(\frac{\partial\ell}{\partial x_1}, \frac{\partial\ell}{\partial x_2}, \cdots, \frac{\partial\ell}{\partial x_n})^\text{T}, \frac{\partial\ell}{\partial \vect{h}}=(\frac{\partial\ell}{\partial h_1}, \frac{\partial\ell}{\partial h_2}, \cdots, \frac{\partial\ell}{\partial h_m})^\text{T}$ \\
and $\frac{\partial\ell}{\partial \mat{W}}=(\frac{\partial\ell}{\partial W_{ij}})_{1\le i\le m, 1\le j\le n}$, then
\begin{align}
\frac{\partial\ell}{\partial \mat{W}}=\frac{\partial\ell}{\partial \vect{h}}\vect{x}^\text{T},\quad
\frac{\partial\ell}{\partial \vect{x}}=\mat{W}^\text{T}\frac{\partial \ell}{\partial \vect{h}},\quad
\frac{\partial\ell}{\partial \vect{h}}=\frac{\partial\ell}{\partial \vect{z}}\odot{\sigma'(\vect{h})}
\end{align}

In the proposed unified sparse backpropagation framework, the sparsifying function (Definition~\ref{def:sparsifing_function}) is utilized to sparsify the gradient $\frac{\partial\ell}{\partial\vect{h}}$ propagated from the next layer and propagates them through the gradient computation graph according to the chain rule. Note that $\frac{\partial\ell}{\partial\vect{h}}$ is also an estimated gradient passed from the next layer. The gradient estimations are finally performed as follows:
\begin{align}
\frac{\partial\ell}{\partial \vect{h}}\gets\frac{\partial\ell}{\partial \vect{z}}\odot{\sigma'(\vect{h})}, \quad
\frac{\partial\ell}{\partial \mat{W}}\gets S_{\mathbb{I}_k}\left(\frac{\partial\ell}{\partial \vect{h}}\right)\vect{x}^\text{T}, \quad
\frac{\partial\ell}{\partial \vect{x}}\gets\mat{W}^\text{T}S_{\mathbb{I}_k}\left(\frac{\partial \ell}{\partial \vect{h}}\right)
\end{align}
in other words,
\begin{align}
\vect{g}^{\vect{w}}=flatten(\vect{g^y}\vect{x}^\text{T}),\quad
\vect{g^x}=\mat{W}^\text{T}\vect{g^y},\quad
\vect{g^y}=S_{\mathbb{I}_k}({\vect{g^z}\odot{\sigma'(\vect{h})}})
\end{align}

We introduce a lemma:
\begin{lem}
\label{lemma.meprop.1}
For a linear layer trained with SBP, the sparse ratio of the sparsifying function in SBP is $r$. Denote $\vect{w}=flatten(\mat{W})$ .If $\ell=\ell(\vect{w}, (\vect{x}, \vect{y}))$ is the loss of MLP trained with SBP on this input instance and the input of this layer is $\vect{x}$ which satisfies $\|\vect{x}\| \ne 0$, we use SBP to estimate $\partial\ell/\partial  \vect{w}$ and $\partial\ell/\partial \vect{x}$. suppose $\mat{W}\mat{W}^\text{T}$ is a positive definite matrix, the eigenvalue decomposition of $\mat{W}\mat{W}^\text{T}$ is $\mat{W}\mat{W}^\text{T}=\mat{P}\mat{\Sigma} \mat{P}^\text{T}\in \mathbb{R}^{m\times m}$, $\mat{\Sigma}=\text{diag}\{s_1, s_2, \cdots\, s_m \}$ ($s_i>0$) and $\mat{P}$ is an orthogonal matrix. We define ${s_\text{min}}=\min\limits_{i}s_i$, ${s_\text{max}}=\max\limits_{i}s_i$ and ${\sigma'_\text{min}}=\min\limits_{i}\sigma'(h_i)$, ${\sigma'_\text{max}}=\max\limits_{i}\sigma'(h_i)$. It is easy to verify that $s_\text{min}>0$ and $\sigma'_\text{min}>0$ because $\mat{W}\mat{W}^\text{T}$ is a positive definite matrix and $\sigma$ is increasing. If $\rho_1\ge\frac{s_\text{max}}{s_\text{min}}\ge 1$, $\rho_2\ge (\frac{\sigma'_\text{max}}{\sigma'_\text{min}})^2\ge 1$, $\rho_1\cos\delta(\vect{h})+1-\rho_1>0$ and $\rho_2\cos\delta(\vect{z})+1-\rho_2>0$, then we have
$$\delta(\vect{x})\le\arccos\big(\rho_1\cos\delta(\vect{h})+1-\rho_1\big)$$
and
$$\delta(\vect{w})=\delta(\vect{h})\le\arccos\sqrt{r}+\arccos\big(\rho_2\cos\delta(\vect{z})+1-\rho_2\big)$$
\end{lem}

\subsection{MLP Trained with SBP}
Consider a MLP trained with SBP, it is a $N$-layer multi-layer perception (MLP), every layer except the last layer is a linear layer with SBP. $\vect{x}^{(1)}\in\mathbb{R}^{n_1}$ is the input of the MLP, $\vect{x}^{(N+1)}\in\mathbb{R}^{n_{N+1}}$ is the output of the MLP. The $i$-th layer of MLP is defined as
\begin{align}
\vect{h}^{(i)}=\mat{W}^{(i)}\vect{x}^{(i)}, \quad
\vect{x}^{(i+1)}=\sigma_i(\vect{h}^{(i+1)})
\end{align}
where $\vect{x}^{(i)}\in\mathbb{R}^{n_i}$, $\mat{W}^{(i)}\in\mathbb{R}^{n_{i+1}\times n_i}$ and $n_{i+1}\ge n_i$, $\sigma_i$ is an increasing pointwise activation function of layer $i$ ($i<N$). Note that the last layer is not a linear layer trained with SBP. Therefore, $\sigma_N$ need not to be an increasing pointwise activation function. It can be softmax function, which is not a pointwise activation function.

Assume $\vect{w}$ is the parameter vector of MLP defined as
$$\vect{w}=[{\vect{w}^{(1)}}^\text{T}, {\vect{w}^{(2)}}^\text{T}, \cdots, {\vect{w}^{(N)}}^\text{T}]^\text{T}\in\mathbb{R}^{n_{total}},\quad
n_{total}=n_1n_2+n_2n_3+\cdots+n_{N-1}n_N+n_Nn_{N+1}$$
where $\vect{w}^{(i)}=flatten(\mat{W}^{(i)})\in 
\mathbb{R}^{n_in_{i+1}}$.

We use the condition number to measure how sensitive the output is to perturbations in the input data and to roundoff errors made during the solution process. Define condition number of matrix $\mat{A}$ as $\text{cond}(\mat{A})=\|\mat{A}\|\|\mat{A}^{-1}\|$, when we adopts the spectral norm $\|\mat{A}\|=\|\mat{A}\|_2$, then $\text{cond}(\mat{A})=\frac{\lambda_\text{max}}{\lambda_\text{min}}$,
where $\lambda_\text{max}$ and $\lambda_\text{min}$ are the maximum and minimum singular value of $\mat{A}$ respectively. 

If the condition number is small, we say the matrix is well-posed and otherwise ill-posed. If a matrix is singular, then its condition number is infinite, it is very ill-posed. 

For a MLP trained with SBP, we assume that it is \textbf{$\rho$-well-posed} if there exist $\rho_1>1$, $\rho_2>1$ in any layer $i$ and any time step $t$ such that
\begin{align}
\rho_1\ge(\text{cond}(\mat{W}_t^{(i)}))^2,\quad
\rho_2\ge \big(\text{cond}(\text{diag}[\sigma'_i(\vect{h}^{(i+1)}_t)])\big)^2,\quad
\rho=\rho_1\rho_2
\end{align}
here for a $n$-dim vector $\vect{v}=[v_1, v_2, \cdots, v_n]^\text{T}$, we define $\text{diag}[\vect{v}]=\text{diag}\{v_1, v_2, \cdots, v_n\}$. 

We introduce a lemma here to ensure that the gradient estimation angle of the parameter vector can be arbitrarily small for an input instance with its label as input in MLP trained with SBP.

\begin{lem}
\label{thm2}
For a MLP trained with SBP, for any input instance $\vect{x}^{(1)}=\vect{x}$ with its label $\vect{y}$ which satisfies $\|\vect{x}\|\ne 0$. Assume $\vect{w}$ is the parameter vector. If the MLP is $\rho$-well-posed, then for any $\theta\in (0, {\pi}/{2})$, there exsits $r\in({1}/{\rho^2}, 1)$ such that if we set the sparse ratio of every sparsifying function in SBP as $r$, we can get $\vect{g}^{\vect{w};(\vect{x}, \vect{y})}$, an estimation of $\nabla\ell\big(\vect{w}; (\vect{x}, \vect{y})\big)$ to make gradient estimation angle satisfy $\delta(\vect{w})<\theta$.
\end{lem}

\subsection{Review of the Term "In Probability"}
A sequence of random variables $X_n$ converges to a random variable $X$ in probability if for any $\epsilon>0$
\begin{align}
\lim\limits_{n\to\infty}\mathrm{P}(|X_n-X|\ge\epsilon)=0
\end{align}

We introduce a lemma here
\begin{lem}
\label{lemma.p.1}
For a sequence of random variables $X_n$, when $n\to\infty$, if $\mathrm{Var}(X_n)\to 0$ and $\mathrm{E}(X_n)\to a$, then $X_n$ converges to $a$ in probability.
\end{lem}

\section{Proofs}

\subsection{Proofs of Theorem~\ref{thm1}}

\begin{proof}
According to the Ineq.(\ref{mu.2}),
\begin{align}
\nabla\ell(\vect{w})\bm{\cdot}(\vect{w}^*-\vect{w}_t)\le -{\mu\over 2}\|\vect{w}_t-\vect{w}^*\|^2+\ell(\vect{w}^*)-\ell(\vect{w}_t) \label{lemma.gd.1.1}
\end{align}

According to Ineq.(\ref{equ-min}),
\begin{align}
\ell(\vect{w}^*)-\ell(\vect{w}_t)\le-{1\over 2L}\|\nabla \ell(\vect{w}_t)\|^2
\label{lemma.gd.1.2}
\end{align}

Combining Ineq.(\ref{lemma.gd.1.1}) with Ineq.(\ref{lemma.gd.1.2}), we have
\begin{align}
\nabla\ell(\vect{w}_t)\bm{\cdot}(\vect{w}^*-\vect{w}_t) & \le -{\mu\over 2}\|\vect{w}_t-\vect{w}^*\|^2-{1\over 2L}\|\nabla \ell(\vect{w}_t)\|^2 \\
& \le  -\sqrt{\mu \over L}\|\vect{w}_t-\vect{w}^*\|\|\nabla \ell(\vect{w}_t)\| \quad \text{(Basic Inequality)}
\end{align}
in other words,
\begin{align}
\cos\angle\langle\nabla\ell(\vect{w}_t),(\vect{w}_t-\vect{w}^*)\rangle
=\frac{(\vect{w}_t-\vect{w}^*)\bm{\cdot}\nabla \ell(\vect{w}_t)}{\|\vect{w}_t-\vect{w}^*\|\|\nabla \ell(\vect{w}_t)\|}
\ge\sqrt{\frac{\mu}{L}}
=\cos\phi(\ell)
\end{align}
we have
\begin{align}
\angle\langle\nabla\ell(\vect{w}_t),\vect{w}-\vect{w}^*\rangle\le\phi(\ell)
\end{align}

According to Lemma~\ref{lemma.vector.1}
\begin{align}
\angle\langle\vect{g}^\vect{w}, \vect{w}-\vect{w}^*\rangle
\le\angle\langle\vect{g}^\vect{w},\nabla\ell(\vect{w})\rangle+\angle\langle\nabla\ell(\vect{w}), \vect{w}-\vect{w}^*\rangle
\le\delta(\vect{w})+\phi(\ell)
\le\theta
\end{align}

Therefore
\begin{align}
\|\vect{w}_{t+1}-\vect{w}^*\|^2 & = \|\vect{w}-\eta \vect{g}_{t}^{\vect{w}}-\vect{w}^*\|^2\\ 
& = \|\vect{w}-\vect{w}^*\|^2+\|\eta \vect{g}_{t}^{\vect{w}}\|^2-2\eta\vect{g}_{t}^{\vect{w}}\bm{\cdot}(\vect{w}-\vect{w}^*) \\  
& = \|\vect{w}-\vect{w}^*\|^2+\|\eta \vect{g}_{t}^{\vect{w}}\|^2-2\eta\cos\angle\langle\vect{g}_{t}^{\vect{w}}, \vect{w}-\vect{w}^*\rangle\|\vect{g}_{t}^{\vect{w}}\|\| \vect{w}-\vect{w}^*\| \\  
& \le \|\vect{w}-\vect{w}^*\|^2+\eta^2 \|\vect{g}_{t}^{\vect{w}}\|^2-2\eta\cos\theta \|\vect{g}_{t}^{\vect{w}}\|\| \vect{w}-\vect{w}^*\|
\end{align}

By setting $\eta={\cos\theta\|\vect{w}-\vect{w}^*\|\over \|\vect{g}_{t}^{\vect{w}}\|}$, we have
\begin{align}
\|\vect{w}_{t+1}-\vect{w}^*\|  & \le \sqrt{\|\vect{w}-\vect{w}^*\|^2+\eta^2 \|\vect{g}_{t}^{\vect{w}}\|^2-2\eta\cos\theta \|\vect{g}_{t}^{\vect{w}}\|\| \vect{w}-\vect{w}^*\|} \\
&=\sqrt{\|\vect{w}-\vect{w}^*\|^2+\frac{\cos^2\theta\|\vect{w}-\vect{w}^*\|^2}{\|\vect{g}_{t}^{\vect{w}}\|^2} \|\vect{g}_{t}^{\vect{w}}\|^2-2\cos^2\theta\| \vect{w}-\vect{w}^*\|^2} \\
&=\sqrt{(1-\cos^2\theta)}\|\vect{w}-\vect{w}^*\| 
=\sin\theta\|\vect{w}-\vect{w}^*\|
\end{align}

Define $a(\theta) = \left. 1 \middle/ \log{\frac{1}{\sin\theta}} \right.$, where $a(\theta)>0$. Then
\begin{align}
\frac{1}{\sin\theta} = \exp{1\over a(\theta)},\quad
\|\vect{w}_{t}-\vect{w}^*\|={1\over \sin\theta}\|\vect{w}_{t+1}-\vect{w}^*\|=\exp{1\over a(\theta)}\|\vect{w}_{t+1}-\vect{w}^*\|
\end{align}
in other words,
\begin{align}
\log\|\vect{w}_{t}-\vect{w}^*\|\ge {1\over a(\theta)}+\log\|\vect{w}_{t+1}-\vect{w}^*\|
\end{align}

We have
\begin{align}
\log\|\vect{w}_{0}-\vect{w}^*\| & \ge  {1\over a(\theta)}+\log\|\vect{w}_{1}-\vect{w}^*\| 
\ge {2\over a(\theta)}+
\ge \cdots 
\ge {T\over a(\theta)}+ \log\|\vect{w}_{T}-\vect{w}^*\|
\end{align}

To ensure that $\|\vect{w}_{T}-\vect{w}^*\|\le\epsilon$, we just have to ensure that $
\log\|\vect{w}_{T}-\vect{w}^*\|+{T\over a(\theta)}\le\log\|\vect{w}_{0}-\vect{w}^*\|\le \log\epsilon+{T\over a(\theta)}$. In other words, we just have to ensure that $T\ge T(\epsilon)= a(\theta)\log{\|\vect{w}_0-\vect{w}^*\| \over \epsilon}$.
Therefore, $\forall \epsilon\in(0, \|\vect{w}_0-\vect{w}^*\|), \exists T(\epsilon)=\log{\|\vect{w}_0-\vect{w}^*\| \over \epsilon} \big/ \log{\frac{1}{\sin\theta}}$ s.t.
\begin{equation}
\|\vect{w}_T-\vect{w}^*\| \le \epsilon \ (\forall T\ge T(\epsilon))
\end{equation}
\end{proof}

\subsection{Proof of Theorem~\ref{thm3}}
\begin{proof}
Suppose $\mathcal{D}={(\vect{x}_i, \vect{y}_i)}_{i=1}^{n}$, where $n$ is the number of data instances. Define
\begin{align}
\vect{u}_i=\ell\big(\vect{w}; (\vect{x}_i, \vect{y}_i)\big),\quad
\vect{v}_i=\vect{g}^{\vect{w}; (\vect{x}_i, \vect{y}_i)}
\end{align}
then
\begin{align}
\nabla\ell(\vect{w}; \mathcal{D})=\frac{1}{|\mathcal{D}|}\sum\limits_{(\vect{x}, \vect{y})\in \mathcal{\mathcal{D}}}\nabla\ell\big(\vect{w}; (\vect{x}, \vect{y})\big),\quad
\vect{g}^{\vect{w}}=\frac{1}{|\mathcal{\mathcal{D}}|}\sum\limits_{(\vect{x}, \vect{y})\in \mathcal{\mathcal{D}}}\vect{g}^{\vect{w};(\vect{x}, \vect{y})} 
\label{equ:estimate}
\end{align}

We introduce a lemma here and we will prove it later. $\vect{u}_i$ and $\vect{v}_i$ in the lemma is defined above.
\begin{lem}
Dataset $\mathcal{D}$ has $n$ independent and identically distributed (i.i.d.) data instances. Suppose for any $\theta\in(0, \frac{\pi}{2})$ and any $(\vect{x}_i, \vect{y}_i)\in \mathcal{D}$ we can find $r\in({1}/{\rho^2}, 1)$ such that if we set the sparse ratio of every sparsifying function in SBP as $r$, then $\angle\langle\vect{u}_i,\vect{v}_i\rangle<\theta$ and $\|\vect{u}_i\|=\|\vect{v}_i\|$. Then for any $\epsilon\in(0,\frac{\pi}{2})$, there exists $r\in({1}/{\rho^2}, 1)$  such that when we set the sparse ratio of every sparsifying function in SBP as $r$, then $\lim\limits_{|\mathcal{D}|\to\infty}\mathrm{P}(\delta(\vect{w})<\epsilon)=1$.
\label{lem:almost_everywhere}
\end{lem}

According to Lemma~\ref{thm2}, for any $\theta\in(0, \frac{\pi}{2})$ and any $(\vect{x}_i, \vect{y}_i)\in \mathcal{D}$ we can find $r\in({1}/{\rho^2}, 1)$ such that if we set the sparse ratio of every sparsifying function in SBP as $r$, then $\angle\langle\vect{u}_i,\vect{v}_i\rangle<\theta$ and $\|\vect{u}_i\|=\|\vect{v}_i\|$ (Eq.(\ref{ui=vi})). We also have a large enough dataset $\mathcal{D}$, which has $n$ independent and identically distributed (i.i.d.) data instances. The condition of Lemma~\ref{lem:almost_everywhere} is satisfied.

Therefore, for $\theta\in(0,\frac{\pi}{2})$, there exists $r\in({1}/{\rho^2}, 1)$  such that when $n\to \infty$ and we set the sparse ratio of every sparsifying function in SBP as $r$, $\lim\limits_{|\mathcal{D}|\to\infty}\mathrm{P}(\delta(\vect{w})<\epsilon)=1$.
\end{proof}


\subsection{Proofs of Lemmas}

\begin{proof}[\textbf{Proof of Lemma~\ref{lemma.vector.1}}]
Without loss of generality, we assume $\|\vect{a}\|=\|\vect{b}\|=\|\vect{c}\|=1$.

Define $\vect{a}_1=\vect{a}-(\vect{a}\bm{\cdot}\vect{c})\vect{c}$ and $\vect{b}_1=\vect{b}-(\vect{b}\bm{\cdot}\vect{c})\vect{c}$. We have
\begin{align}
\|\vect{a_1}\|^2&=\|\vect{a}-(\vect{a}\bm{\cdot}\vect{c})\vect{c}\|^2
=\|\vect{a}\|^2-2\big(\vect{a}\bm{\cdot}((\vect{a}\bm{\cdot}\vect{c})\vect{c})\big)+\|(\vect{a}\bm{\cdot}\vect{c})\vect{c}\|^2 \\
&=\|\vect{a}\|^2-2\vect{a}\bm{\cdot}(\cos\angle\langle\vect{a},\vect{c}\rangle\vect{c})+(\cos\angle\langle\vect{a},\vect{c}\rangle)^2\|\vect{c}\|^2\\
&=1-\cos^2\angle\langle\vect{a},\vect{c}\rangle =\sin^2\angle\langle\vect{a},\vect{c}\rangle\\
\vect{c}\bm{\cdot}\vect{a}_1&=\vect{c}\bm{\cdot}\big(\vect{a}-(\vect{a}\bm{\cdot}\vect{c})\vect{c}\big)
(\vect{c}\bm{\cdot}\vect{a})-(\vect{c}\bm{\cdot}\vect{a})\|\vect{c}\|^2=0
\end{align}

For $\vect{b_1}$, similarly $\|\vect{b_1}\|^2=\sin^2\angle\langle\vect{b},\vect{c}\rangle, 
\vect{c}\bm{\cdot}\vect{b}_1=0$. Therefore,
\begin{align}
\cos\angle\langle\vect{a},\vect{b}\rangle & = \vect{a}\bm{\cdot}\vect{b} =\big(\cos\angle\langle\vect{a},\vect{c}\rangle\vect{c}+\vect{a}_1\big) \bm{\cdot} \big(\cos\angle\langle\vect{b},\vect{c}\rangle\vect{c}+\vect{b}_1)\\
&=\cos\angle\langle\vect{a},\vect{c}\rangle \cos\angle\langle\vect{b},\vect{c}\rangle + \vect{a}_1\bm{\cdot}\vect{b}_1 \\
&\ge \cos\angle\langle\vect{a},\vect{c}\rangle \cos\angle\langle\vect{b},\vect{c}\rangle - \sin\angle\langle\vect{a},\vect{c}\rangle \sin\angle\langle\vect{b},\vect{c}\rangle \\
&= \cos\big(\angle\langle\vect{a},\vect{c}\rangle + \angle\langle\vect{b},\vect{c}\rangle\big)
\end{align}
In other words, $\angle\langle\vect{a},\vect{b}\rangle \le \angle\langle\vect{a},\vect{c}\rangle + \angle\langle\vect{b},\vect{c}\rangle$.
\end{proof}

\begin{proof}[\textbf{Proof of Lemma~\ref{lemma.vector.2}}]
Without loss of generality, we assume $\|\vect{u}\|=\|\vect{v}\|=1$.

$\mat{A}\mat{A}^\text{T}=\mat{P}\mat{\Sigma} \mat{P}^\text{T}$ where $\mat{\Sigma}=\text{diag}\{\sigma_1, \sigma_2, \cdots\, \sigma_m \}$, according to singular value decomposition (SVD), we have $\mat{A}=\mat{P}\mat{D}\mat{Q}^\text{T}$, where $\mat{P}$ and $\mat{Q}$ are orthogonal and $\mat{D}=\text{diag}\{\sqrt{\sigma_1}, \cdots, \sqrt{\sigma_n}\}$.

We can see $\mat{D}^\text{T}=\mat{D}$ and for any vector $\vect{x}=(x_1, x_2, ..., x_m)^\text{T}\in\mathbb{R}^m$
\begin{align}
\|\mat{D}x\|^2
&=\sum\limits_{i=1}^m(\sqrt{\sigma_i}x_i)^2
\le\sigma_\text{max}\sum\limits_{i=1}^mx_i^2
=\sigma_\text{max}\|\vect{x}\|^2 \label{Dmin}\\
\|\mat{D}x\|^2
&=\sum\limits_{i=1}^m(\sqrt{\sigma_i}x_i)^2
\ge\sigma_\text{min}\sum\limits_{i=1}^mx_i^2
=\sigma_\text{min}\|\vect{x}\|^2 \label{Dmax}
\end{align}

We define $\vect{a}=\mat{P}^\text{T}\vect{u}$ and $\vect{b}=\mat{P}^\text{T}\vect{v}$, we have
\begin{align}
\|\vect{a}\|^2
=\vect{a}^\text{T}\vect{a}
=\vect{u}^\text{T}\mat{P}\mat{P}^\text{T}\vect{u}
=\vect{u}^\text{T}\vect{u}
=\|\vect{u}\|^2
=1 \label{adota}
\end{align}
and similarly
\begin{align}
\|\vect{b}\|^2
=\|\vect{v}\|^2
=1 \label{bdotb}
\end{align}
we have
\begin{align}
\cos\angle\langle\vect{a}, \vect{b}\rangle
=\frac{\vect{a}^\text{T}\vect{b}}{\|\vect{a}\|\|\vect{b}\|}
=\frac{\vect{u}^\text{T}\mat{P}\mat{P}^\text{T}\vect{v}}{\|\vect{a}\|\|\vect{b}\|}
=\frac{\vect{u}^\text{T}\vect{v}}{\|\vect{u}\|\|\vect{v}\|}
=\cos\angle\langle\vect{u}, \vect{v}\rangle \label{cosab}
\end{align}
similarly
\begin{align}
\cos\angle\langle \mat{Q}^\text{T}\mat{D}\vect{a}, \mat{Q}^\text{T}\mat{D}\vect{b}\rangle
=\frac{(\mat{D}\vect{a})^\text{T}\mat{Q}\mat{Q}^\text{T}(\mat{D}\vect{b})}{\|\mat{Q}^\text{T}\mat{D}\vect{a}\|\|\mat{Q}^\text{T}\mat{D}\vect{b}\|}
=\frac{(\mat{D}\vect{a})^\text{T}(\mat{D}\vect{b})}{\|\mat{D}\vect{a}\|\|\mat{D}\vect{b}\|}
=\cos\angle\langle \mat{D}\vect{a}, \mat{D}\vect{b}\rangle \label{cosab}
\end{align}

Then $\mat{A}^\text{T}\vect{u}=\mat{Q}^\text{T}\mat{D}\vect{a}$ and $\mat{A}^\text{T}\vect{v}=\mat{Q}^\text{T}\mat{D}\vect{b}$.
Consider
\begin{align}
\|\vect{a}-\vect{b}\|^2
&=\|\vect{a}\|^2+\|\vect{b}\|^2-2\vect{a}\bm{\cdot}\vect{b}
=2(1-\cos\angle\langle\vect{a}, \vect{b}\rangle) \label{a-b} \\
\|\mat{D}(\vect{a}-\vect{b})\|^2&=\|\mat{D}\vect{a}\|^2+\|\mat{D}\vect{b}\|^2-2(\mat{D}\vect{a})\bm{\cdot}(\mat{D}\vect{b}) \label{Da-Db}
\end{align}

According to Eq.(\ref{Dmin}), Eq.(\ref{Dmax}), Eq.(\ref{adota}), Eq.(\ref{bdotb}), Eq.(\ref{cosab}), Eq.(\ref{a-b}) and Eq.(\ref{Da-Db}), we have
\begin{align}
\cos\angle\langle \mat{D}\vect{a}, \mat{D}\vect{b}\rangle&=\frac{(\mat{D}\vect{a})\bm{\cdot}(\mat{D}\vect{b})}{\|\mat{D}\vect{a}\|\|\mat{D}\vect{b}\|} \\
&=\frac{\|\mat{D}\vect{a}\|^2+\|\mat{D}\vect{b}\|^2-\|\mat{D}(\vect{a}-\vect{b})\|^2}{2\|\mat{D}\vect{a}\|\|\mat{D}\vect{b}\|}\\
&=\frac{\|\mat{D}\vect{a}\|^2+\|\mat{D}\vect{b}\|^2}{2\|\mat{D}\vect{a}\|\|\mat{D}\vect{b}\|}-\frac{\|\mat{D}(\vect{a}-\vect{b})\|^2}{2\|\mat{D}\vect{a}\|\|\mat{D}\vect{b}\|}\\
&\ge 1-\frac{\|\mat{D}(\vect{a}-\vect{b})\|^2}{2\|\mat{D}\vect{a}\|\|\mat{D}\vect{b}\|} \quad \text{(Basic Inequality)}\\
&\ge 1-\frac{2\sigma_\text{max}(1-\cos\angle\langle\vect{a}, \vect{b}\rangle)}{2\sigma_\text{min}}\\
&\ge 1-\rho(1-\cos\angle\langle \vect{a}, \vect{b}\rangle)\\
&=\rho\cos\angle\langle\vect{u}, \vect{v}\rangle+1-\rho
\end{align}
and
\begin{align}
\cos\angle\langle \mat{A}^\text{T}\vect{u}, \mat{A}^\text{T}\vect{v}\rangle
=\cos\angle\langle \mat{Q}^\text{T}\mat{D}\vect{a}, \mat{Q}^\text{T}\mat{D}\vect{b}\rangle
=\cos\angle\langle \mat{D}\vect{a}, \mat{D}\vect{b}\rangle
\end{align}
In other words,
\begin{align}
\cos\angle\langle \mat{A}^\text{T}\vect{u}, \mat{A}^\text{T}\vect{v}\rangle\ge\rho\cos\angle\langle\vect{u}, \vect{v}\rangle+1-\rho
\end{align}

\end{proof}

\begin{proof}[\textbf{Proof of Lemma~\ref{lemma.meprop.1}}]

First, let's consider $\delta(\vect{x})$.

According to Lemma~\ref{lemma.vector.2}
\begin{align}
\cos\delta(\vect{x})=\cos\angle\langle\vect{g^x}, \frac{\partial\ell}{\partial\vect{x}}\rangle 
=\cos\angle\langle \mat{W}^\text{T}\vect{g^y}, \mat{W}^\text{T}\frac{\partial\ell}{\partial\vect{h}}\rangle \\
\ge \rho_1\cos\angle\langle \vect{g^y}, \frac{\partial\ell}{\partial\vect{h}}\rangle+1-\rho_1 
= \rho_1\cos\delta(\vect{h})+1-\rho_1
\end{align}
In other words, 
\begin{align}
\delta(\vect{x})\le\arccos\big(\cos\delta(\vect{h})+\rho_1-1\big) \label{deltaxdeltay}
\end{align}

Then, let's consider $\delta(\vect{h})$.

According to Lemma~\ref{lemma.vector.1}
\begin{align}
\delta(\vect{h})&=\angle\langle\vect{g^y}, \frac{\partial\ell}{\partial\vect{h}}\rangle 
=\angle\langle\textbf{S}\big(\sigma'(\vect{h})\odot\vect{g^z}\big), \sigma'(\vect{h})\odot\frac{\partial\ell}{\partial\vect{z}}\rangle \\
&\le \angle\langle\textbf{S}\big(\sigma'(\vect{h})\odot\vect{g^z}\big), \sigma'(\vect{h})\odot\vect{g^z} \rangle+\angle\langle \sigma'(\vect{h})\odot\vect{g^z}, \sigma'(\vect{h})\odot\frac{\partial\ell}{\partial\vect{z}}\rangle \\
&\le \arccos\sqrt{r}+\angle\langle \sigma'(\vect{h})\odot\vect{g^z}, \sigma'(\vect{h})\odot\frac{\partial\ell}{\partial\vect{z}}\rangle \quad \text{(Ineq.(\ref{costopk}))} \label{deltay}
\end{align}

Define $\mat{A}=\text{diag}\{\sigma'(h_1),\sigma'(h_2),\cdots,\sigma'(h_m)\}$,\\
then $\mat{A}\mat{A}^\text{T}=\text{diag}\{\sigma'(h_1)^2,\sigma'(h_2)^2,\cdots,\sigma'(h_m)^2\}$, according to Lemma~\ref{lemma.vector.2}
\begin{align}
\cos\angle\langle \sigma'(\vect{h})\odot\vect{g^z}, \sigma'(\vect{h})\odot\frac{\partial\ell}{\partial\vect{z}}\rangle
= \cos\angle\langle \mat{A}^\text{T}\vect{g^z}, \mat{A}^\text{T}\frac{\partial\ell}{\partial\vect{z}}\rangle 
\ge\rho_2\cos\angle\langle \vect{g^z}, \frac{\partial\ell}{\partial\vect{z}}\rangle+1-\rho_2
\end{align}
In other words, $\angle\langle \sigma'(\vect{h})\odot\vect{g^z}, \sigma'(\vect{h})\odot\frac{\partial\ell}{\partial\vect{z}}\rangle\le\arccos\big(\rho_2\cos\angle\langle \vect{g^z}, \frac{\partial\ell}{\partial\vect{z}}\rangle+1-\rho_2\big)$.

Combined with Ineq.(\ref{deltay}), we have 
\begin{align}
\delta(\vect{h}) &\le \arccos\sqrt{r}+\angle\langle \sigma'(\vect{h})\odot\vect{g^z}, \sigma'(\vect{h})\odot\frac{\partial\ell}{\partial\vect{z}}\rangle \\
&\le\arccos\sqrt{r}+\arccos\big(\rho_2\cos\angle\langle \vect{g^z}, \frac{\partial\ell}{\partial\vect{z}}\rangle+1-\rho_2\big) \\
&=\arccos\sqrt{r}+\arccos\big(\rho_2\cos\delta(\vect{z})+1-\rho_2\big)
\end{align}

Finally, let's consider $\delta(\vect{w})=\delta(flatten(\mat{W}))$.

Without loss of generality, we assume
\begin{align}
|\frac{\partial\ell}{\partial h_1}|\ge|\frac{\partial\ell}{\partial h_2}|\ge...\ge|\frac{\partial\ell}{\partial h_m}|
\label{cond.absy}
\end{align}

On one hand,
\begin{align}
\|\vect{g}^{\vect{w}}\|^2
=\sum\limits_{i=1}^{n}\sum\limits_{j=1}^{m}\left(x_i\vect{g}^\vect{h}_j\right)^2
=\big(\sum\limits_{i=1}^{n}x_i^2\big)\big(\sum\limits_{j=1}^{m}(\vect{g}^\vect{h}_j)^2\big) 
= \|\vect{x}\|^2 \|\vect{g}^\vect{h}\|^2\label{gdotg}
\end{align}

On the other hand,
\begin{align}
\|\frac{\partial\ell}{\partial \vect{w}}\|^2
=\sum\limits_{i=1}^{n}\sum\limits_{j=1}^{m}\left(x_i\frac{\partial\ell}{\partial h_j}\right)^2 
=\left(\sum\limits_{i=1}^{n}x_i^2\right)\left(\sum\limits_{j=1}^{m}(\frac{\partial\ell}{\partial h_j})^2\right) 
= \|\vect{x}\|^2 \|\frac{\partial\ell}{\partial \vect{h}}\|^2\label{wdotw}
\end{align}

Consider
\begin{align}
\vect{g}^{\vect{w}}\bm{\cdot}\frac{\partial\ell}{\partial \vect{w}}
= \sum\limits_{i=1}^{n}\sum\limits_{j=1}^{m}\left(x_i\vect{g}^\vect{h}_j\right)\left(x_i\frac{\partial\ell}{\partial h_j}\right)
=\left(\sum\limits_{i=1}^{n}x_i^2\right)\left(\sum\limits_{j=1}^{m}(\vect{g}^\vect{h}_j\frac{\partial\ell}{\partial h_j})\right)
= \|\vect{x}\|^2(\vect{g^y}\bm{\cdot}\frac{\partial\ell}{\partial\vect{h}})
\end{align}
combined with Eq.(\ref{gdotg}) and Eq.(\ref{wdotw})
\begin{align}
\cos\delta(\vect{w})=\frac{\vect{g}^{\vect{w}}\bm{\cdot}\frac{\partial\ell}{\partial \vect{w}}}{\|\vect{g}^{\vect{w}}\|\|\frac{\partial\ell}{\partial \vect{w}}\|} 
=\frac{\|\vect{x}\|^2(\vect{g^y}\bm{\cdot}\frac{\partial\ell}{\partial\vect{h}})}{\|\vect{x}\|^2\|\vect{g^y}\|\|\frac{\partial\ell}{\partial\vect{h}}\|} 
=\cos\angle\langle\vect{g^y}, \frac{\partial\ell}{\partial\vect{h}}\rangle 
=\cos\delta(\vect{h})
\end{align}
In other words, $\delta(\vect{w})=\delta(flatten(\mat{W}))=\delta(\vect{h})$.
\end{proof}

\begin{proof}[\textbf{Proof of Lemma~\ref{thm2}}]

For $\vect{w}=[{\vect{w}^{(1)}}^\text{T}, {\vect{w}^{(2)}}^\text{T}, \cdots, {\vect{w}^{(N)}}^\text{T}]^\text{T}\in\mathbb{R}^{n_{total}}, n_{total}=n_1n_2+n_2n_3+\cdots+n_{N-1}n_N+n_Nn_{N+1}$, if we define the estimated gradient $\vect{g}$ as $\vect{g}=[\lambda^{(1)}(\vect{g}^{\vect{w}^{(1)}})^\text{T}, \lambda^{(2)}(\vect{g}^{\vect{w}^{(2)}})^\text{T}, \cdots, \lambda^{(N)}(\vect{g}^{\vect{w}^{(N)}})^\text{T}]^\text{T}\in\mathbb{R}^{n_{total}}$. We use $\vect{g}^{\vect{w};(\vect{x}, \vect{y})}=\vect{g}$ to estimate $\nabla\ell\big(\vect{w}; (\vect{x}, \vect{y})\big)$ and the estimated angle is
\begin{align}
\delta(\vect{w})=\angle\langle\vect{g}, \frac{\partial\ell\big(\vect{w};(\vect{x}, \vect{y})\big)}{\partial\vect{w}}\rangle
\end{align}

We choose $\lambda^{(i)}={\|\frac{\partial\ell}{\partial\vect{w}^{(i)}}\|}\big/{\|\vect{g}^{\vect{w}^{(i)}}\|}$, then we have
\begin{align}
(\lambda^{(i)}\vect{g}^{\vect{w}^{(i)}})\bm{\cdot}\frac{\partial\ell}{\partial\vect{w}^{(i)}}
&=\lambda^{(i)}\|\vect{g}^{\vect{w}^{(i)}}\|\|\frac{\partial\ell}{\partial\vect{w}^{(i)}}\|\cos\delta(\vect{w}^{(i)})
=\|\frac{\partial\ell}{\partial\vect{w}^{(i)}}\|^2\cos\delta(\vect{w}^{(i)})\\
\|\vect{g}\|^2
&=\sum\limits_{i=1}^N\|\lambda^{(i)}\vect{g}^{\vect{w}^{(i)}}\|^2
=\sum\limits_{i=1}^N\|\frac{\partial\ell}{\partial\vect{w}^{(i)}}\|^2
=\|\frac{\partial\ell}{\partial\vect{w}}\|^2 \label{ui=vi}
\end{align}

Suppose $\delta=\max\limits_{i}\delta(\vect{w}^{(i)})$, then we have
\begin{align}
\cos\delta(\vect{w})&=\cos\angle\langle\vect{g}, \frac{\partial\ell}{\partial\vect{w}}\rangle 
=\frac{\vect{g}\bm{\cdot}\frac{\partial\ell}{\partial\vect{w}}}{\|\vect{g}\|\|\frac{\partial\ell}{\partial\vect{w}}\|} 
=\frac{\sum\limits_{i=1}^N(\lambda^{(i)}\vect{g}^{\vect{w}^{(i)}})\bm{\cdot}\frac{\partial\ell}{\partial\vect{w}^{(i)}}}{\|\frac{\partial\ell}{\partial\vect{w}}\|^2} \\
&=\frac{\sum\limits_{i=1}^N\|\frac{\partial\ell}{\partial\vect{w}^{(i)}}\|^2\cos\delta(\vect{w}^{(i)})}{\|\frac{\partial\ell}{\partial\vect{w}}\|^2} 
\ge\frac{\sum\limits_{i=1}^N\|\frac{\partial\ell}{\partial\vect{w}^{(i)}}\|^2\cos\delta}{\|\frac{\partial\ell}{\partial\vect{w}}\|^2} 
= \cos\delta
\end{align}
In other words, $\delta(\vect{w})\le\delta$.

We will prove that there exists $r\in(\frac{1}{\rho^2}, 1)$ to ensure $\delta<\theta$.

For a $\rho$-well-posed $N$-layer MLP trained with SBP, there exist $\rho_1>1, \rho_2>1$ satisfying
\begin{align}
\rho_1\ge(\text{cond}(\mat{W}^{(i)}))^2,\quad
\rho_2\ge \big(\text{cond}(\text{diag}[\sigma'_i(\vect{h}^{(i+1)})])\big)^2,\quad
\rho=\rho_1\rho_2
\end{align}
therefore, denote $\mat{W}=\mat{W}^{(i)}$ and $\vect{h}^{(i+1)}=[h_1, h_2, \cdots, h_i]^\text{T}$, if the eigenvalue decomposition of $\mat{W}\mat{W}^\text{T}$ is $\mat{W}\mat{W}^\text{T}=\mat{P}\mat{\Sigma} \mat{P}^\text{T}\in \mathbb{R}^{m\times m}$ ($s_i>0$), $\mat{\Sigma}=\text{diag}\{s_1, s_2, \cdots\, s_m \}$ and $\mat{P}$ is an orthogonal matrix (${s_\text{min}}=\min\limits_{i}s_i$, ${s_\text{max}}=\max\limits_{i}s_i$) and ${\sigma'_\text{min}}=\min\limits_{i}\sigma'(h_i)$, ${\sigma'_\text{max}}=\max\limits_{i}\sigma'(h_i)$. (It is easy to verify that $s_\text{min}>0$ and $\sigma'_\text{min}>0$ because $\mat{W}\mat{W}^\text{T}$ is a positive definite matrix and $\sigma$ is increasing.) We have
\begin{align}
\rho_1\ge(\text{cond}(\mat{W}^{(i)}))^2=\frac{s_\text{max}}{s_\text{min}} ,\quad
\rho_2\ge (\frac{\sigma'_\text{max}}{\sigma'_\text{min}})^2 ,\quad
\rho=\rho_1\rho_2
\end{align}

Note that $\rho_1>1, \rho_2>1$ satisfying the conditions in Lemma~\ref{lemma.meprop.1} for every linear layer with SBP and $\rho=\rho_1\rho_2>1$

Define $\alpha_i=\cos\delta(\vect{w}^{(i)})$, note that the last layer is not with SBP, therefore $\alpha_N=1$. For $i<N$, if $\rho\alpha_{i+1}+1-\rho=\rho_1\rho_2\alpha_{i+1}+1-\rho_1\rho_2>0$, we have
\begin{align}
&\rho_1\cos\delta(\vect{h}^{(i+1)})+1-\rho_1
=\rho_1\alpha_{i+1}+1-\rho_1\\
&\ge\rho_1\rho_2\alpha_{i+1}+1-\rho_1\rho_2
>0\\
&\rho_2\cos\delta(\vect{x}^{(i+1)})+1-\rho_2
\ge\rho_2\big(\rho_1\cos\delta(\vect{h}^{(i+1)})+1-\rho_1\big)+1-\rho_2\\
&=\rho_1\rho_2\alpha_{i+1}+1-\rho_1\rho_2
>0 \quad \text{(Ineq.(\ref{deltaxdeltay}))}
\end{align}
which are conditions of Lemma~\ref{lemma.meprop.1} exactly, according to Lemma~\ref{lemma.meprop.1}
\begin{align}
\delta(\vect{x}^{(i)})
\le\arccos\big(\rho_1\cos\delta(\vect{h}^{(i)})+1-\rho_1\big)
=\arccos\big(\rho_1\alpha_i+1-\rho_1\big) \label{cosdeltax}\\
\arccos\alpha_i
=\delta(\vect{w}^{(i)})
=\delta(\vect{h}^{(i)})
\le\arccos\sqrt{r}+\arccos\big(\rho_2\cos\delta(\vect{x}^{(i+1)})+1-\rho_2\big) \label{cosdeltay}
\end{align}
In other words,
\begin{align}
\alpha_i&\ge\cos\big(\arccos\sqrt{r}+\arccos\big(\rho_2\cos\delta(\vect{x}^{(i+1)})+1-\rho_2\big)\big) \quad \text{(Ineq.(\ref{cosdeltay}))}\\
&\ge \cos\big(\arccos\sqrt{r}+\arccos(\rho_2(\rho_1\alpha_{i+1}+1-\rho_1)+1-\rho_2)\big) \quad \text{(Ineq.(\ref{cosdeltax}))}\\
&= \cos\big(\arccos\sqrt{r}+\arccos\big(\rho_1\rho_2\alpha_{i+1}+1-\rho_1\rho_2\big)\big)\\
&= \cos\big(\arccos\sqrt{r}+\arccos\big(\rho\alpha_{i+1}+1-\rho\big)\big)\\
&= \sqrt{r}(\rho\alpha_{i+1}+1-\rho)-\sqrt{1-r}\sqrt{1-(\rho\alpha_{i+1}+1-\rho)^2} \label{alpha_i_alpha_i+1}
\end{align}

Define $\beta_N=0\ge 1-\alpha_N$ and for $1\le i< N$
\begin{align}
1-\beta_i=\sqrt{r}(\rho(1-\beta_{i+1})+1-\rho)-\sqrt{1-r}\sqrt{1-(\rho(1-\beta_{i+1})+1-\rho)^2} \label{beta_i_beta_i+1}
\end{align}

Assume $\beta_1<\frac{1}{\rho}<1$ first, then
\begin{align}
1-\beta_1&=\sqrt{r}(\rho(1-\beta_{2})+1-\rho)-\sqrt{1-r}\sqrt{1-(\rho(1-\beta_{2})+1-\rho)^2} \\
&=\sqrt{r}\rho(1-\beta_{2})-\sqrt{r}(\rho-1)-\sqrt{1-r}\sqrt{1-(\rho(1-\beta_{2})+1-\rho)^2} \\
&<1-\beta_2 \quad \text{(here we can see $\beta_2<1$)}
\end{align}
therefore $\beta_2<\beta_1<\frac{1}{\rho}<1$.

Similarly if $\beta_i<\frac{1}{\rho}<1$
\begin{align}
1-\beta_i&=\sqrt{r}(\rho(1-\beta_{i+1})+1-\rho)-\sqrt{1-r}\sqrt{1-(\rho(1-\beta_{i+1})+1-\rho)^2} \\
&=\sqrt{r}\rho(1-\beta_{i+1})-\sqrt{r}(\rho-1)-\sqrt{1-r}\sqrt{1-(\rho(1-\beta_{i+1})+1-\rho)^2} \\
&<1-\beta_{i+1} \quad \text{(here we can see $\beta_{i+1}<1$)}
\end{align}
therefore $\beta_{i+1}<\beta_{i}<\cdots<\beta_2<\beta_1<\frac{1}{\rho}<1$. In other words, $\beta_N<\beta_{N-1}<\cdots<\beta_2<\beta_1<\frac{1}{\rho}<1$. 

If $\beta_{i+1}\ge1-\alpha_{i+1}$, we have $\rho\alpha_{i+1}+1-\rho=1-\rho(1-\alpha_{i+1})>0$ (because $\frac{1}{\rho}>\beta_1>\beta_{i+1}$), which is the condition for  Ineq.(\ref{alpha_i_alpha_i+1}). According to  Ineq.(\ref{alpha_i_alpha_i+1}) and  Ineq.(\ref{beta_i_beta_i+1})
\begin{align}
1-\beta_i&=\sqrt{r}(\rho(1-\beta_{i+1})+1-\rho)-\sqrt{1-r}\sqrt{1-(\rho(1-\beta_{i+1})+1-\rho)^2} \\
&\le \sqrt{r}(\rho\alpha_{i+1}+1-\rho)-\sqrt{1-r}\sqrt{1-(\rho\alpha_{i+1}+1-\rho)^2} 
\le \alpha_i
\end{align}
In other words, $\beta_{i}\ge1-\alpha_{i}$. Note $\beta_N\ge1-\alpha_N$, therefore $\beta_N<\beta_{N-1}<\cdots<\beta_2<\beta_1<\frac{1}{\rho}<1,\quad
\beta_{i}\ge1-\alpha_{i}$.

In order to ensure $\delta<\theta$ under the assumption $\beta_1<\frac{1}{\rho}$
\begin{align}
\cos\delta=\cos\max\limits_i\delta(\vect{w}_t^{(i)}) 
=\cos\min\limits_i \arccos \alpha_i 
=\max\limits_i \alpha_i \ge \max\limits_i (1-\beta_i)
=1-\beta_1
\end{align}
we just need to ensure $\beta_1<\min(\frac{1}{\rho}, 1-\cos\theta)
\label{ensure_beta}$. According to Eq.(\ref{beta_i_beta_i+1})
\begin{align}
\beta_i&=1-\sqrt{r}(1-\rho\beta_{i+1})+\sqrt{1-r}\sqrt{1-(1-\rho\beta_{i+1})^2} \\
&=\rho\sqrt{r}\beta_{i+1}+1-\sqrt{r}+\sqrt{1-r}\sqrt{1-(1-\rho\beta_{i+1})^2} \\
&<\rho\sqrt{r}\beta_{i+1}+1-\sqrt{r}+\sqrt{1-r}
\end{align}

Denote $a=\rho\sqrt{r}>1, b=1-\sqrt{r}+\sqrt{1-r}>0$, then
\begin{align}
\beta_1<a\beta_2+b
<a(a\beta_3+b)+b
=a^2\beta_3+(a+1)b
<a^2(a\beta_4+b)+(a+1)b \\
<\cdots
<a^{N-1}\beta_N+(a^{N-2}+\cdots+a+1)b 
=(a^{N-2}+\cdots+a+1)b
=\frac{a^{N-1}-1}{a-1}b
\end{align}

Therefore, we just need to ensure
\begin{align}
\beta_1<\frac{a^{N-1}-1}{a-1}b<\min(\frac{1}{\rho}, 1-\cos\theta) \label{ensure_beta_1}
\end{align}

Denote $f(r)=\min(\frac{1}{\rho}, 1-\cos\theta)-\frac{a^{N-1}-1}{a-1}b$, where $a=\rho\sqrt{r}>1, b=1-\sqrt{r}+\sqrt{1-r}>0$. To ensure $\delta(\vect{w})<\theta$, we just need to ensure $f(r)>0$. We have
\begin{align}
&\lim\limits_{r\to 1^-}b
=\lim\limits_{r\to 1^-}(1-\sqrt{r}+\sqrt{1-r})
=0 \\
&\lim\limits_{r\to 1^-}\frac{a^{N-1}-1}{a-1}
=\lim\limits_{r\to 1^-}\frac{(N-1)a^{N-2}}{1}
=N-1 \quad \text{(L'Hospital rule)}\\
&\lim\limits_{r\to 1^-}f(r)
=\min(\frac{1}{\rho}, 1-\cos\theta)-(N-1)\times0
=\min(\frac{1}{\rho}, 1-\cos\theta)
>0
\end{align}
because $f(r)$ is a continuous function of $r$, therefore there exists $r\in(\frac{1}{\rho^2}, 1)$ such that $f(r)>0$.

\end{proof}

\begin{proof}[\textbf{Proof of Lemma~\ref{lemma.p.1}}]
According to Chebyshev's Ineq.
\begin{align}
\mathrm{P}(|X_n-\mathrm{E}(X_n)|<\epsilon)> 1-\frac{\mathrm{Var}(X_n)}{\epsilon^2}
\end{align}
In other words,
\begin{align}
\mathrm{P}(|X_n-\mathrm{E}(X_n)|<\epsilon)\to 1 (n\to\infty)
\end{align}

Because $\mathrm{E}(X_n)\to a$ in probability, we have the following in probability
\begin{align}
\mathrm{P}(|\mathrm{E}(X_n)-a|<\epsilon)\to 1 (n\to\infty)
\end{align}

For event $A, B$, we have 
\begin{align}
\mathrm{P}(A \text{ and } B) 
= 1-\mathrm{P}(\text{not } A \text{ or } \text{not } B) 
\ge 1-\mathrm{P}(\text{not } A)-\mathrm{P}(\text{not } B) 
=\mathrm{P}(A)+\mathrm{P}(B)-1
\end{align}
combined with
\begin{align}
\mathrm{P}(|X_n-a|<2\epsilon)
&\ge\mathrm{P}(|\mathrm{E}(X_n)-a|<\epsilon\text{ and } |X_n-\mathrm{E}(X_n)|<\epsilon)\\
&\ge \mathrm{P}(|\mathrm{E}(X_n)-a|<\epsilon)+\mathrm{P}( |X_n-\mathrm{E}(X_n)|<\epsilon)-1
\end{align}

Therefore
\begin{align}
\mathrm{P}(|X_n-a|<2\epsilon)\to 1 (n\to\infty)
\end{align}
\end{proof}

\begin{proof}[\textbf{Proof of Lemma~\ref{lem:almost_everywhere}}]
For any $\theta$, we can choose $r=r(\theta)$ to let $\angle\langle\vect{u}_i,\vect{v}_i\rangle<\theta$, we define such $r$ as $r(\theta)$. To ensure $\delta(\vect{w}_t)<\epsilon$, we just need to ensure $\cos\delta(\vect{w}_t)>\cos\epsilon$. 

Define 
\begin{align}
\vect{\bar u}=\frac{1}{n}\sum\limits_{i=1}^{n}\vect{u}_i=\nabla\ell(\vect{w}_t; \mathcal{D}),\quad
\vect{\bar v}=\frac{1}{n}\sum\limits_{i=1}^{n}\vect{v}_i=\vect{g}_t^\vect{w}
\end{align}
then
\begin{align}
\cos\delta(\vect{w}_t)
=\cos\angle\langle\vect{g}_t^\vect{w}, \nabla\ell(\vect{w}_t; \mathcal{D})\rangle
=\cos\angle\langle\vect{\bar u}, \vect{\bar v}\rangle
=\frac{1}{n\|\vect{\bar v}\|}\sum\limits_{i=1}^n\|\vect{v}_i\|\cos\angle\langle\vect{v}_i, \vect{\bar u}\rangle \label{cosdelta1}
\end{align}

According to Lemma~\ref{lemma.vector.1}, $\angle\langle\vect{v}_i, \vect{\bar u}\rangle\le\angle\langle\vect{v}_i, \vect{u}_i\rangle+\angle\langle\vect{u}_i, \vect{\bar u}\rangle<\theta+\angle\langle\vect{u}_i, \vect{\bar u}\rangle$. Because $\theta$ can be arbitrarily small, $\theta+\angle\langle\vect{u}_i, \vect{\bar u}\rangle<\pi$ can hold. Define 
\begin{align}
a_i=\|\vect{v}_i\|\cos\angle\langle\vect{u}_i, \vect{\bar u}\rangle,\quad
b_i=\|\vect{v}_i\|\sin\angle\langle\vect{u}_i, \vect{\bar u}\rangle
\end{align}
According to Minkowski Ineq.
\begin{align}
(\sum\limits_{i=1}^na_i)^2+(\sum\limits_{i=1}^nb_i)^2\le \big(\sum\limits_{i=1}^n\sqrt{a_i^2+b_i^2}\big)^2
\end{align}
Define$\beta=\sum\limits_{i=1}^n\|\vect{u}_i\|/(n\|\vect{\bar u}\|)$, then,
\begin{align}
\sum\limits_{i=1}^na_i
=\sum\limits_{i=1}^n\|\vect{v}_i\|\cos\angle\langle\vect{u}_i, \vect{\bar u}\rangle
=\sum\limits_{i=1}^n\|\vect{u}_i\|\cos\angle\langle\vect{u}_i, \vect{\bar u}\rangle
=\frac{1}{\|\vect{\bar u}\|}\sum\limits_{i=1}^n\vect{u}_i\bm{\cdot}\vect{\bar u}
=n\|\vect{\bar u}\|\\
\sum\limits_{i=1}^nb_i
\le\sqrt{\big(\sum\limits_{i=1}^n\sqrt{a_i^2+b_i^2}\big)^2-(\sum\limits_{i=1}^na_i)^2}
=\sqrt{(\sum\limits_{i=1}^n\|\vect{u}_i\|)^2-(n\|\vect{\bar u}\|)^2}
=n\|\vect{\bar u}\|\sqrt{\beta^2-1}
\end{align}

Combined with Eq.(\ref{cosdelta1}), then
\begin{align}
\cos\delta(\vect{w}_t)
&=\frac{1}{n\|\vect{\bar v}\|}\sum\limits_{i=1}^n\|\vect{v}_i\|\cos\angle\langle\vect{v}_i, \vect{\bar u}\rangle
>\frac{1}{n\|\vect{\bar v}\|}\sum\limits_{i=1}^n\|\vect{v}_i\|\cos(\angle\langle\vect{u}_i, \vect{\bar u}\rangle+\theta)\\
&=\frac{1}{n\|\vect{\bar v}\|}\sum\limits_{i=1}^n\big\|\vect{v}_i\|(\cos\angle\langle\vect{u}_i, \vect{\bar u}\rangle\cos\theta-\sin\angle\langle\vect{u}_i, \vect{\bar u}\rangle\sin\theta\big)\\
&=\frac{1}{n\|\vect{\bar v}\|}\sum\limits_{i=1}^n(a_i\cos\theta-b_i\sin\theta)
=\frac{1}{n\|\vect{\bar v}\|}(\sum\limits_{i=1}^na_i\cos\theta-\sum\limits_{i=1}^nb_i\sin\theta)\\
&\ge\frac{1}{n\|\vect{\bar v}\|}(n\|\vect{\bar u}\|\cos\theta-n\|\vect{\bar u}\|\sqrt{\beta^2-1}\sin\theta)
=\frac{\|\vect{\bar u}\|}{\|\vect{\bar v}\|}(\cos\theta-\sqrt{\beta^2-1}\sin\theta) \label{cosdelta2}
\end{align}

Consider
\begin{align}
\|\vect{\bar u}-\vect{\bar v}\|
&=\frac{1}{n} \|\sum\limits_{i=1}^n(\vect{u}_i-\vect{v}_i)\|
\le\frac{1}{n}\sum\limits_{i=1}^n\|\vect{u}_i-\vect{v}_i\|\\
&=\frac{1}{n}\sum\limits_{i=1}^n\sqrt{\|\vect{u}_i\|^2+\|\vect{v}_i\|^2-2\cos\angle\langle\vect{u}_i, \vect{v}_i\rangle}
<\frac{1}{n}\sum\limits_{i=1}^n\sqrt{\|\vect{u}_i\|^2(2-2\cos\theta)}\\
&=\frac{1}{n}\sum\limits_{i=1}^n2\|\vect{u}_i\|\sin\frac{\theta}{2}
=2\beta\|\vect{\bar u}\|\sin\frac{\theta}{2}
\end{align}
In other words,
\begin{align}
\frac{\|\vect{\bar v}\|}{\|\vect{\bar u}\|}
=\frac{\|\vect{\bar v}-\vect{\bar u}+\vect{\bar u}\|}{\|\vect{\bar u}\|}
\le\frac{\|\vect{\bar v}-\vect{\bar u}\|+\|\vect{\bar u}\|}{\|\vect{\bar u}\|}
<1+2\beta\sin\frac{\theta}{2}
\end{align}

Because $\theta$ can be arbitrarily small, $\cos\theta-\sqrt{\beta^2-1}\sin\theta>0$ can hold. Combined with Eq.(\ref{cosdelta2}), then
\begin{align}
\cos\delta(\vect{w}_t)
>\frac{\|\vect{\bar u}\|}{\|\vect{\bar v}\|}(\cos\theta-\sqrt{\beta^2-1}\sin\theta)
>\frac{1}{1+2\beta\sin\frac{\theta}{2}}(\cos\theta-\sqrt{\beta^2-1}\sin\theta) \label{cosdelta3}
\end{align}

We define $f(\theta)=\frac{1}{1+2\beta\sin\frac{\theta}{2}}(\cos\theta-\sqrt{\beta^2-1}\sin\theta)$, where $\beta=\sum\limits_{i=1}^n\|\vect{u}_i\|/(n\|\vect{\bar u}\|)$.

Assume $\|\vect{u}_i\|$ is i.i.d., $\mathrm{Var}\|\vect{u}_i\|$ and $\mathrm{E}\|\vect{u}_i\|$ are finite, and $\|\mathrm{E}\vect{u}_i\|>0$. (It is reasonable because the data instances are i.i.d. and we may assume the gradients' norm is bounded, and also if $\|\mathrm{E}\vect{u}_i\|=0$, the network already converges to the global minimum). 

Note that if $A$ and $B$ are independent, $\mathrm{Var}(A+B)=\mathrm{Var}(A)+\mathrm{Var}(B), \mathrm{E}(AB)=\mathrm{E}(A)\mathrm{E}(B)$. We have
\begin{align}
\mathrm{Var}(\frac{\sum\limits_{i=1}^n\|\vect{u}_i\|}{n})
&=\frac{1}{n^2}\sum\limits_{i=1}^n\mathrm{Var}\|\vect{u}_i\|
=\frac{1}{n}\mathrm{Var}\|\vect{u}_i\|
\to 0\quad (n\to\infty)\\
\mathrm{Var}(\vect{\bar u}^{(j)})
&=\mathrm{Var}(\frac{\sum\limits_{i=1}^n\vect{u}_i^{(j)}}{n})
=\frac{1}{n}\mathrm{Var}(\vect{u}_i^{(j)})
\to 0\quad (n\to\infty)
\end{align}
where $\vect{u}_i^{(j)}, \vect{\bar u}^{(j)}$ represent the $j$-th dim of the vector.

According to Lemma~\ref{lemma.p.1} and $\mathrm{E}({\sum\limits_{i=1}^n\|\vect{u}_i\|}/{n})=\mathrm{E}(\|\vect{u}_i\|)$, when $n\to\infty$ (here we consider convergence in probability),
\begin{align}
\frac{\sum\limits_{i=1}^n\|\vect{u}_i\|}{n}\to\mathrm{E}\|\vect{u}_i\|,\quad
\vect{\bar u}^{(j)}\to \mathrm{E}\vect{u}_i^{(j)},\quad
\|\vect{\bar u}\|\to \|\mathrm{E}\vect{u}_i\|,\quad
\beta\to\frac{\mathrm{E}\|\vect{u}_i\|}{\|\mathrm{E}\vect{u}_i\|}
\end{align}

Note that we assume $\mathrm{Var}\|\vect{u}_i\|$ and $\mathrm{E}\|\vect{u}_i\|$ are finite and $\|\mathrm{E}\vect{u}_i\|>0$. Therefore, there exists $\beta_1$ such that $\beta_1>{\mathrm{E}\|\vect{u}_i\|}/{\|\mathrm{E}\vect{u}_i\|}$ holds in every time step. Therefore when $n$ is large enough,
\begin{align}
\mathrm{P}(\beta<\beta_1)\ge\mathrm{P}(|\beta-\frac{\mathrm{E}\|\vect{u}_i\|}{\|\mathrm{E}\vect{u}_i\|}|<\beta_1-\frac{\mathrm{E}\|\vect{u}_i\|}{\|\mathrm{E}\vect{u}_i\|}) \to 1
\end{align}

When $\beta<\beta_1$, we have
\begin{align}
f(\theta)=
\frac{1}{1+2\beta\sin\frac{\theta}{2}}(\cos\theta-\sqrt{\beta^2-1}\sin\theta)
>\frac{1}{1+2\beta_1\sin\frac{\theta}{2}}(\cos\theta-\sqrt{\beta^2_1-1}\sin\theta)
\end{align}

To ensure $\delta(\vect{w}_t)<\epsilon$, we just need to ensure $f(\theta)>\cos\epsilon$, consider
\begin{align}
\lim\limits_{\theta\to0}\frac{1}{1+2\beta_1\sin\frac{\theta}{2}}(\cos\theta-\sqrt{\beta^2_1-1}\sin\theta)\to 1
\end{align}
In other words, for any $\epsilon$, there exists $\theta$ and $r$ such that if we set the sparse ratio $r=r(\theta)$ then $\cos(\delta(\vect{w}_t))<\epsilon$ holds when $\beta<\beta_1$. Therefore when $n$ is large enough,
\begin{align}
\mathrm{P}(\cos(\delta(\vect{w}_t))<\epsilon)\ge\mathrm{P}(\beta<\beta_1) \to 1
\end{align}

To conclude,
\begin{align}
\lim\limits_{n\to\infty}\mathrm{P}(\cos(\delta(\vect{w}_t))<\epsilon) = 1
\end{align}

\end{proof}

\section{Statistical test}
In this section, statistical tests are conducted on MNIST dataset for MSBP and SBP under different settings of $k$ and $\gamma$. We may assume that the accuracies of SBP and MSBP obey two normal distributions $N(\mu_1, \sigma_1)$ and $N(\mu_2, \sigma_2)$ respectively. Repeating times of both SBP and MSBP are $n=20$.

1) First, to test whether MSBP improves the performance, Student t-tests are conducted: Null hypothesis $H_0: \mu_1\ge\mu_2$, alternative hypothesis $H_a: \mu_1<\mu_2$. $t$-value $\approx1.7$ when $p=0.05$ and the degree of freedom $df=2(n-1)=38$. Results for different settings are shown in Table~\ref{ttest}. 

\begin{table*}[!h]
\caption{Results of $t$-values under different settings.}
\centering
\footnotesize
\renewcommand\tabcolsep{5pt}
\begin{tabular}{|c|ccccccccc|}
\hline
\diagbox{$k$}{$t$-value}{$\gamma$} & $0.1$ & $0.2$ & $0.3$ & $0.4$ & $0.5$ & $0.6$ & $0.7$ & $0.8$ & $0.9$\\
\hline
$5$ & $16.2$ & $14.6$ & $18.0$ & $17.0$ & $19.3$ & $18.8$ & $18.2$ & $18.5$ & $16.4$ \\
$10$ & $11.1$ & $12.0$ & $11.2$ & $12.6$ & $10.7$ & $11.2$ & $11.8$ & $11.6$ & $11.1$ \\
$20$ & $6.3$ & $5.2$ & $5.8$  & $5.8$ & $5.1$ & $6.2$ & $5.1$ & $6.4$ & $5.8$ \\
\hline
\end{tabular}
\label{ttest}
\end{table*}

For all settings of $k$ and $\gamma$, $t$-values$\ge 1.7$, that is, MSBP improves the performance of SBP statistically significantly ($p<0.05$).

2) Then, to test whether MSBP improves the stability, F-tests are conducted: Null hypothesis $H_0: \sigma_1\le\sigma_2$, alternative hypothesis $H_a: \sigma_1>\sigma_2$. $F$-value $\approx 2.1$ when $p=0.05$ and the degrees of freedom of numerator and denominator are both $df=n-1=19$. Results for different settings are shown in Table~\ref{ftest}.

\begin{table*}[!h]
\caption{Results of $F$-values under different settings.}
\centering
\footnotesize
\renewcommand\tabcolsep{5pt}
\begin{tabular}{|c|ccccccccc|}
\hline
\diagbox{$k$}{$F$-value}{$\gamma$} & $0.1$ & $0.2$ & $0.3$ & $0.4$ & $0.5$ & $0.6$ & $0.7$ & $0.8$ & $0.9$\\
\hline
$5$ & $4.9$ & $\textbf{1.9}$ & $4.0$ & $2.6$ & $7.3$ & $3.8$ & $4.4$ & $3.4$ & $4.0$ \\
$10$ & $2.9$ & $5.5$ & $2.2$ & $2.1$ & $2.2$ & $2.8$ & $3.5$ & $3.2$ & $3.6$ \\
$20$ & $6.2$ & $2.9$ & $5.2$  & $3.6$ & $4.3$ & $5.7$ & $2.7$ & $3.0$ & $7.0$ \\
\hline
\end{tabular}
\label{ftest}
\end{table*}

For nearly all settings of $k$ and $\gamma$ (except the setting where $k=5, \gamma=0.2$, which is bold in Table~\ref{ftest}), $F$-values$\ge 2.1$, that is, MSBP improves the stability of SBP statistically significantly ($p<0.05$).

To conclude, the proposed MSBP method improves both the performance and stability of traditional SBP statistically significantly ($p<0.05$) for nearly all settings of $k$ and $\gamma$.

\end{document}